\documentclass[lettersize,journal]{IEEEtran}
\usepackage{amsmath,amsfonts}
\usepackage{algorithmic}
\usepackage{algorithm}
\usepackage{array}
\usepackage[caption=false,font=normalsize,labelfont=sf,textfont=sf]{subfig}
\usepackage{textcomp}
\usepackage{stfloats}
\usepackage{url}
\usepackage{verbatim}
\usepackage{graphicx}
\usepackage{cite}

\usepackage{microtype}
\usepackage{graphicx}
\usepackage{subcaption}
\usepackage{booktabs}
\usepackage{algorithm} 
\usepackage{algorithmic}

\usepackage{multirow}
\usepackage{float}
\usepackage{siunitx}
\usepackage{silence}
\usepackage{stfloats}
\usepackage[table]{xcolor}
\usepackage[textsize=tiny]{todonotes}
\usepackage{hyperref}
\hypersetup{
    colorlinks=true,
    linkcolor=black,
    citecolor=black,
    urlcolor=black
}
\usepackage[capitalize,noabbrev]{cleveref}
\begin{document}

\title{AffectOmni: RL-Verifiable People-Centric Grounded Affective Reasoning for Social and Art-Related Scenes}

\author{%
Yibo~Wang,
Rui~Yang,
Jisheng~Dang,
Bimei~Wang,
Yitao~Wu,
Pengfei~Cao,
Wencan~Zhang,
Hong~Peng,
Bin~Hu,~\IEEEmembership{Fellow,~IEEE,}
and~Tat-Seng~Chua%
\thanks{Yibo Wang, Rui Yang, Jisheng Dang, Bimei Wang, Pengfei Cao, Hong Peng, and Bin Hu are with Lanzhou University, Lanzhou, China.}%
\thanks{Yitao Wu is with Hainan University, Haikou, China.}%
\thanks{Wencan Zhang and Tat-Seng Chua are with the School of Computing, National University of Singapore, Singapore.}%
\thanks{Corresponding authors: Bin Hu, Pengfei Cao, and Wencan Zhang.}%
}



\maketitle

\begin{abstract}
Multimodal large language models (MLLMs) achieve strong performance on VQA and scene understanding, yet affective reasoning remains vulnerable to shortcut behavior. Models may predict correct answers while neglecting people-centric cues such as micro expressions and body language, which weakens traceability and external verification. Prior reinforcement learning approaches mainly reward context or logical coherence without explicitly enforcing attention to human evidence. In addition, LLM as a Judge scoring often suffers from score clustering, which reduces reward discriminability. We propose AffectOmni, a GRPO trained framework for verifiable affective reasoning. AffectOmni introduces People Focus and Temporal Order rewards to encourage people-centric evidence selection and temporally structured reasoning, and it adopts within-group comparative scoring to produce more stable and discriminative reward signals. For verification, a Thinking Summarizer converts free form rationales into executable evidence instructions, which are grounded into pixel level evidence regions via SAM3 to provide an externally auditable interface outside the training loop. Experiments on IntentBench, Daily Omni, and WorldSense show consistent improvements over open source 7B scale baselines, including gains of \textbf{4.66\%} on emotion recognition and \textbf{+14.29\%} on temporally sensitive tasks. Code is available at \url{https://github.com/eliot127825-rgb/AffectOmni_nobody}.
\end{abstract}


\section{Introduction}

\begin{figure}[h]
\centering
\includegraphics[width=\columnwidth]{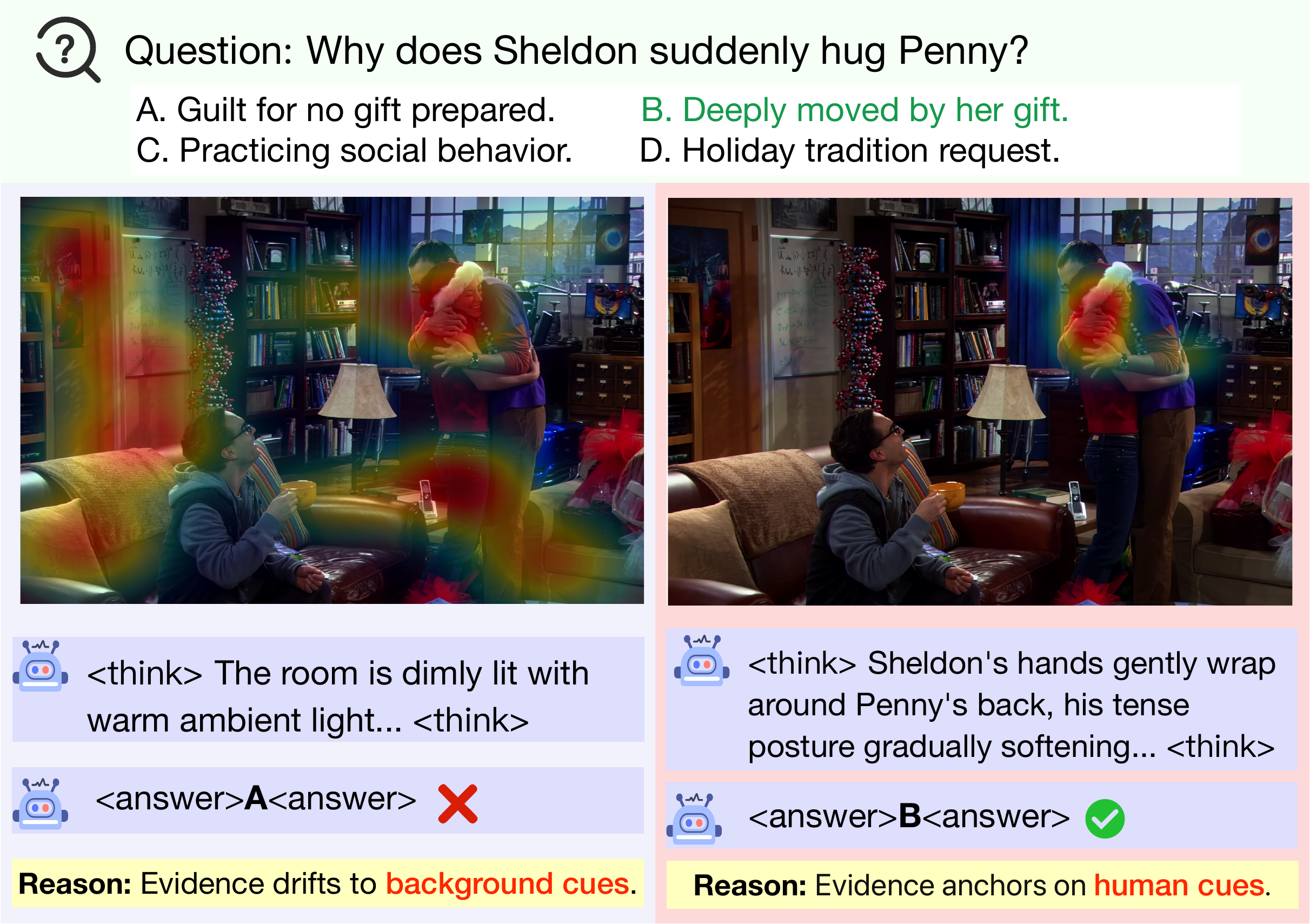}
\caption{\textbf{Shortcut reasoning vs.\ human-centric reasoning.}
Left: without People Focus reward, evidence attribution drifts to background cues, leading to an incorrect answer.
Right: with People Focus reward, evidence anchors on human cues such as hand placement and posture change, enabling evidence-grounded affective inference.
The displayed frame is a representative snapshot from the full video, previous events provide context, while human cues provide discriminative evidence.}
\label{fig:motivation}
\end{figure}

\IEEEPARstart{M}{ultimodal} large language models (MLLMs) have made substantial progress~\cite{achiam2023gpt,team2023gemini,xu2025qwen2}, and reinforcement learning, especially GRPO~\cite{shao2024deepseekmath}, further strengthens their capacity for deep reasoning~\cite{huang2025vision,yuan2025vl}. This capability is widely expected to support high level social cognition tasks involving human emotion and intent~\cite{cambria2017affective,li2020deep,yang2025large}. However, Visionary-R1~\cite{xia2025visionary} and HumanOmniV2~\cite{yang2025humanomniv2} report a pronounced shortcut problem in multimodal reasoning. Models often guess answers from global cues while failing to analyze fine-grained evidence. HumanOmniV2 introduces explicit context modeling and an LLM as a Judge reward~\cite{zheng2023judging}, which partially alleviates limitations in global context understanding. Nevertheless, in affective computing~\cite{zadeh2016multimodal,hu2024recent}, we identify a more fundamental challenge. Even when a model predicts the correct emotion, its reasoning often overlooks people-centric cues such as micro expressions and body language and cannot be verified against external evidence~\cite{reich2023measuring,das2017human}. As shown in Fig.~\ref{fig:motivation}, current models~\cite{yang2025humanomniv2} may rely on background cues rather than people-centric evidence, yielding an untrustworthy reasoning process and sometimes an incorrect answer. This remains a key barrier to deploying affective AI in real world settings~\cite{dong2025integrating,amin2024wide}.

We distill the trustworthiness challenge in affective reasoning into three interrelated questions. 
(1) How can we design fine-grained reward signals that guide the model to learn a people-centric reasoning mode, rather than relying on coarse global cues. 
(2) How can we obtain reward signals with sufficient discriminative power during reinforcement learning to reliably separate evidence grounded reasoning from shortcut reasoning. 
(3) How can we transform long chain reasoning into an executable structured representation and establish a verification mechanism that links reasoning claims to visual evidence.

Existing approaches address these needs only partially. 
For (1), prior rewards such as global context completeness in HumanOmniV2~\cite{yang2025humanomniv2} mainly target holistic context understanding, but they do not explicitly model or encourage key affective cues, including micro expressions such as furrowed brows and teary eye corners, body language such as gestures and posture changes, and temporal dynamics that reflect emotion evolution over time~\cite{li2020deep,lian2023explainable}. 
For (2), most LLM as a Judge mechanisms adopt an independent scoring strategy. Each candidate response is scored in isolation. However, LLMs can exhibit calibration drift and score clustering in absolute scoring settings~\cite{wang2024large,zheng2023judging}, causing reasoning paths of different quality to receive similar scores and yielding weakly discriminative rewards for policy optimization. 
For (3), current models often generate verbose chains of thought of roughly 800 to 1400 characters that contain rich details but remain loosely structured and low density, making them difficult to directly support downstream applications such as counseling assistance and keyframe extraction~\cite{lian2025affectgpt,niu2025rethinking}. Moreover, label only evaluation cannot test whether the model truly attends to the visual regions that are relevant to the emotion judgment~\cite{yu2018mattnet,plummer2015flickr30k}.

Our central thesis is that trustworthy affective reasoning depends not only on answer correctness, but more importantly on a reasoning process that is traceable, reward signals that are discriminative, and outcomes that are externally verifiable. Building on this view, we propose the AffectOmni framework. Our main contributions are as follows.

\begin{itemize}
\item We propose the first reasoning to evidence grounding paradigm for affective reasoning tasks. We compress model generated affective reasoning into executable, structured evidence instructions and interface them with SAM3 to produce pixel level evidence regions on video frames. This provides externally traceable and verifiable evidence anchors for the people, actions, and temporal cues referenced in reasoning.

\item We propose a fine-grained reward mechanism for people-centric reasoning by introducing multi dimensional constraints, including People Focus and Temporal Order, into reinforcement learning. These constraints encourage the model to attend to micro expressions, body language, and human interactions, and to organize reasoning over time.
\item During GRPO training, we introduce a within-group comparative scoring strategy that mitigates calibration drift and score clustering in LLM as a Judge evaluation, thereby providing a more stable and discriminative optimization signal.
\item We achieve state of the art performance on multimodal benchmarks including IntentBench, Daily-Omni, and WorldSense, showing consistent improvements over open source 7B scale baselines, including \textbf{+4.66\%} on emotion recognition and \textbf{+14.29\%} on temporally sensitive tasks.
\end{itemize}

\section{Related Work}

\textbf{Multimodal Affect Understanding.}
Affective computing aims to enable machines to understand human affect, and the field has progressed from rule based classification to deep reasoning~\cite{cambria2017affective,li2020deep}. Early studies focused on unimodal feature extraction and end to end recognition~\cite{zadeh2016multimodal}, followed by explorations of multimodal fusion strategies. Hazarika et al.~\cite{hazarika2020misa} propose modality invariant representation learning, and Hu et al.~\cite{hu2022unimse} build a unified framework for affect and emotion recognition. With the rise of MLLMs, affective computing has entered a new phase. Lian et al.~\cite{lian2023explainable} introduce an explainable multimodal affective reasoning benchmark. Zhou et al.~\cite{zhou2025daily} study an audio visual temporal alignment mechanism. Lian et al.~\cite{lian2025affectgpt} propose AffectGPT for open vocabulary affect understanding. Chain of thought prompting has also been explored for implicit affective reasoning~\cite{lai2025rvisa,yang2025application}. However, existing methods often exhibit shortcut learning. Models may bypass fine-grained cues such as micro expressions and body language and directly predict labels~\cite{xia2025visionary,yang2025humanomniv2}. We introduce a fine-grained reward design centered on people-centric cues to explicitly encourage attention to key affective evidence.

\begin{figure*}[t]
\centering
\includegraphics[width=\textwidth]{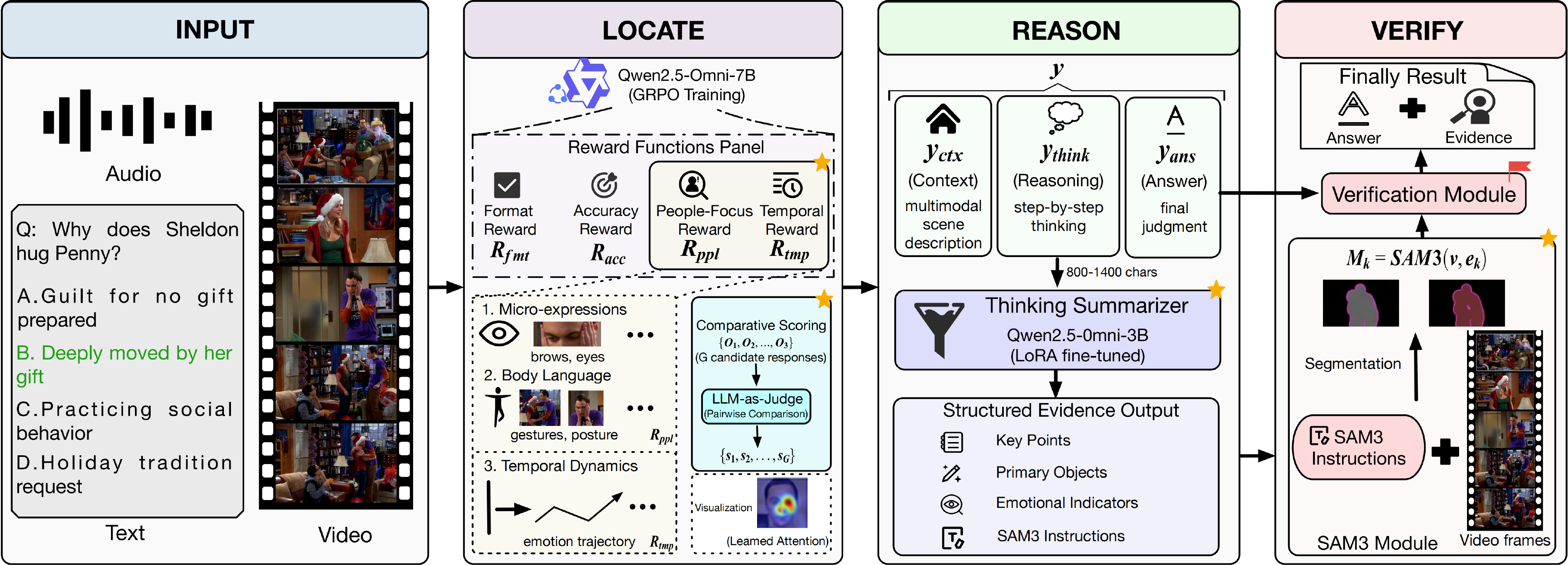}
\caption{\textbf{AffectOmni framework.} The pipeline consists of three stages. (1)~\textbf{LOCATE} applies GRPO training with human-centric rewards
($R_{\mathrm{ppl}}$, $R_{\mathrm{tmp}}$) and comparative scoring to guide
evidence-grounded reasoning. (2)~\textbf{REASON} generates structured output and compresses the reasoning chain into a Minimal Evidence Package (MEP) via a Thinking Summarizer. (3)~\textbf{VERIFY} grounds the MEP to pixel-level masks using SAM3, enabling post-hoc auditing of whether the evidence claims can be localized in the visual input. Stars indicate our key contributions. We further evaluate the verification module in Section~\ref{sec:discussion}.}
\label{fig:framework}
\end{figure*}

\textbf{Reinforcement Learning and Reward Design.}
Reinforcement learning has become a key approach for eliciting the reasoning capability of MLLMs. Huang et al.~\cite{huang2025vision} and Yuan et al.~\cite{yuan2025vl} first apply GRPO to visual reasoning, and subsequent work extends it to video understanding~\cite{wang2025videorft,li2025videochat}, omni modal reasoning~\cite{zhong2025omni,zhu2025active}, and audio visual domains~\cite{xing2025echoink,wang2025sightsound}. To mitigate shortcut learning, recent studies design diverse reward functions. HumanOmniV2~\cite{yang2025humanomniv2} proposes context and logical consistency rewards. Chen et al.~\cite{chen2025unveiling} introduce a step level reward, and Chen et al.~\cite{chen2026grpocare} design a consistency reward. Other variants include format rewards~\cite{zhang2025r1}, perceptual rewards~\cite{guo2025observe}, and hybrid reward strategies~\cite{wang2025skywork,hong2025apo}. Most of these methods follow the LLM as a Judge paradigm, where a language model assigns absolute scores to each generated output in isolation. However, calibration drift and score clustering can arise in absolute scoring, so reasoning paths of different quality may receive similar scores, reducing reward discriminability. We propose within-group comparative scoring, which evaluates multiple candidates jointly and enforces relative ranking to obtain more discriminative reward signals.

\textbf{Visual Grounding and Reasoning Verification.}
Visual grounding aligns language descriptions with regions in images or videos~\cite{yu2018mattnet,plummer2015flickr30k}. Reich et al.~\cite{reich2023measuring} propose faithful and plausible metrics for grounding, and Das et al.~\cite{das2017human} analyze attention discrepancies between humans and deep networks. In medical VQA, studies validate the effectiveness of the localize then answer paradigm~\cite{nguyen2025localizing,chen2022grounding}. For trustworthy MLLMs, grounding can mitigate hallucinations~\cite{favero2024multi} and support multimodal fact checking~\cite{suryavardan2023factify,cekinel2025multimodal}. Yu et al.~\cite{yu2024rlhf} and Sun et al.~\cite{sun2024aligning} use RLHF for behavior alignment, and Zhang et al.~\cite{zhang2025improving} employ reinforcement learning to improve multi image grounding. The SAM family~\cite{ravi2024sam,carion2025sam} enables pixel level video segmentation, providing a technical basis for reasoning verification. Prior work explores visual grounding and post-hoc localization, but it typically treats grounding and affective reasoning as separate components. We integrate a reasoning summarizer with SAM3 based pixel level grounding to provide an externally auditable evidence interface for post-hoc verification of affective reasoning.

\section{Method}

\subsection{Problem Formulation and Framework Overview}

Given a multimodal input $\mathbf{x}$ that includes a video frame sequence $\mathbf{v}$, an audio signal $\mathbf{a}$, and a textual question $q$, our goal is to train a policy model $\pi_\theta$ to generate a structured
reasoning output $o=\langle context, think, answer\rangle$. The context segment describes the multimodal context, the think segment provides step by step reasoning, and the answer segment gives the final decision.

We train the policy under the Group Relative Policy Optimization (GRPO) framework. For each input $\mathbf{x}$, the model samples $G$ candidate responses $\{o_i\}_{i=1}^G$ and updates the policy using within-group relative advantages. The training objective is given in Eq.~(\ref{eq:grpo_obj}) and Eq.~(\ref{eq:grpo_surrogate}).
\begin{equation}
\label{eq:grpo_obj}
\mathcal{J}(\theta)
= \mathbb{E}\Bigg[
\frac{1}{\sum_{i=1}^{G}|o_i|}
\sum_{i=1}^{G}\sum_{t=1}^{|o_i|}
L_{i,t}(\theta)
\Bigg],
\end{equation}
\begin{equation}
\label{eq:grpo_surrogate}
L_{i,t}(\theta)
=
\min\Big(
r_{i,t}\hat{A}_i,\;
\text{clip}(r_{i,t}, 1\pm\varepsilon)\hat{A}_i
\Big),
\end{equation}
where $r_{i,t}=\pi_\theta(o_{i,t}\mid \mathbf{x},o_{i,<t})/\pi_{\theta_{\text{old}}}(o_{i,t}\mid \mathbf{x},o_{i,<t})$ is the importance sampling ratio, and $\hat{A}_i=R_i-\text{mean}(\{R_j\}_{j=1}^G)$ denotes the advantage estimate with a within-group mean baseline. The clipping range is $(1\pm\varepsilon)$.

The proposed AffectOmni framework is illustrated in Fig.~\ref{fig:framework}.
It consists of three complementary components, people-centric reward shaping
for task aligned affective reasoning, within-group comparative scoring for
more discriminative GRPO rewards, and a post-hoc auditing interface that
grounds reasoning claims into visual evidence.
The overall reward is defined as
\begin{equation}
R = R_{\text{fmt}} + R_{\text{acc}} + \lambda_p R_{\text{ppl}} + \lambda_t R_{\text{tmp}},
\label{eq:total_reward}
\end{equation}
where $R_{\text{fmt}}$ and $R_{\text{acc}}$ denote the format reward and the accuracy
reward, $R_{\text{ppl}}$ and $R_{\text{tmp}}$ are our proposed People Focus and
Temporal Order rewards, and $\lambda_p$, $\lambda_t$ are weighting coefficients
used to balance reward components.

\subsection{People-Centric Fine-Grained Reward Shaping}

Existing methods such as HumanOmniV2 adopt a context
level reward that mainly evaluates global context completeness,
but they do not explicitly model key affective cues such as
micro expressions and body language. To address this limitation,
we design two complementary fine-grained reward functions.
These rewards are task-aligned for affective reasoning rather than
universal criteria across domains. For new tasks, the framework
remains unchanged, while only the evaluation criteria in the judge
prompts need to be replaced.

\textbf{People Focus Reward.}
Let $c$ and $h$ denote the context text and the reasoning text generated by the model. We use an LLM as a Judge model to assess whether the reasoning attends to people-centric cues. As shown in Eq.~(\ref{eq:people_reward}),
\begin{equation}
R_{\text{ppl}} = f_{\text{LLM}}(c, h; \mathcal{P}_{\text{ppl}}) \in \{0, 1\},
\label{eq:people_reward}
\end{equation}
where $\mathcal{P}_{\text{ppl}}$ is an evaluation prompt that guides the judge to evaluate three dimensions. The first is facial expression description that covers micro expressions such as furrowed brows and teary eyes. The second is body motion analysis that covers non verbal signals such as gestures and posture changes. The third is interpersonal interaction modeling that captures affective exchange and responses between interlocutors.

\textbf{Temporal Order Reward.}
Affective states evolve over time, so the model should capture temporal changes along the affective trajectory. We therefore introduce a Temporal Order reward
\begin{equation}
R_{\text{tmp}} = f_{\text{LLM}}(c, h; \mathcal{P}_{\text{tmp}}) \in \{0, 1\},
\label{eq:temporal_reward}
\end{equation}
where $\mathcal{P}_{\text{tmp}}$ instructs the judge to evaluate two aspects. The first is the proper use of temporal markers such as initially, then, and finally. The second is temporal coherence of the affective trajectory, such as a gradual transition from surprise to relief.

\textbf{Implementation Details.}
To reduce API overhead, we adopt a joint evaluation scheme. We merge $\mathcal{P}_{\text{ppl}}$ and $\mathcal{P}_{\text{tmp}}$ into a single prompt $\mathcal{P}_{\text{joint}}$ and return two scores in one call. In addition, we apply a causal mask so that $R_{\text{ppl}}$ and $R_{\text{tmp}}$ act only on token positions in the context and think segments, which prevents reward leakage into the answer segment.

\subsection{Within-Group Comparative Scoring}

The independent absolute scoring paradigm based on LLM as a Judge has inherent drawbacks. Large language models can exhibit calibration drift and score clustering in absolute scoring, causing reasoning paths of different quality to receive similar scores and reducing reward discriminability.

\textbf{Comparative Scoring Mechanism.}
We propose a within-group comparative scoring strategy that presents the $G$ candidates for the same prompt, $\{o_1,\ldots,o_G\}$, jointly to the judge model and requires relative comparison rather than absolute scoring. Formally,
\begin{equation}
\{s_1, \ldots, s_G\} = f_{\text{cmp}}(\{o_1, \ldots, o_G\}; \mathcal{P}_{\text{cmp}}),
\label{eq:comparative}
\end{equation}
where $s_i \in [1,10]$ is the relative score for the $i$th candidate. The prompt $\mathcal{P}_{\text{cmp}}$ instructs the judge to perform explicit comparison before assigning scores and to produce a differentiated ranking among candidates.

\textbf{Advantage Computation.}
During training, we compute an overall reward $R_i$ for each candidate as a
weighted sum of multiple reward components
\begin{equation}
R_i=\sum_{k}\lambda_k\, r_i^{(k)},
\label{eq:reward_sum}
\end{equation}
where $r_i^{(k)}$ denotes the $k$th reward component and $\lambda_k$ is the
corresponding weight. Comparative scoring is not a separate reward term; it is
used to estimate judge-based components over the $G$ candidates.

For the $G$ candidates of the same prompt, we compute a within-group mean
baseline and obtain the advantage
\begin{equation}
\hat{A}_i = R_i-\bar{R}_g,\qquad 
\bar{R}_g=\frac{1}{G}\sum_{j=1}^{G}R_j,
\label{eq:advantage}
\end{equation}
and optionally apply within-group standardization
$\hat{A}_i \leftarrow \hat{A}_i/(\sigma_g+\epsilon)$ to improve reward scale
consistency and training stability.

\textbf{Distributed Implementation.}
In multi GPU training, comparative scoring requires aggregating candidates across devices. We use a gather and broadcast scheme. Candidates from all GPUs are gathered to the rank 0 process, which calls the judge model to perform comparative scoring. The resulting scores are then broadcast to all ranks to resume backpropagation.

\subsection{Reasoning to Evidence Grounding Framework}

Long chain reasoning, often around 800 to 1400 characters, can contain rich analytic details, but it is typically low density and loosely structured, which limits its utility for downstream applications and external verification. We propose a reasoning to evidence grounding framework that converts free form reasoning into an executable structured representation and grounds the resulting evidence instructions through visual segmentation. The Thinking Summarizer and SAM3 are used only for post-hoc evidence execution and visualization-based evaluation. The entire verification stage is excluded from GRPO training and is not used as a reward signal. It serves as a decoupled auditing interface rather than an end-to-end optimization loop.

\textbf{Thinking Summarizer.}
We define a mapping from the reasoning output $o$ to a structured summary $z$ as $g_\phi \colon o \mapsto z$, where $z$ contains four fields, key points, primary objects, emotional indicators, and SAM3 instruction. We train $g_\phi$ with a two stage knowledge distillation pipeline. In the first stage, a large API based model produces high quality summaries to serve as training targets. In the second stage, we apply LoRA fine tuning to a lightweight base model with an autoregressive objective
\begin{equation}
\mathcal{L}_{\text{sum}} = -\sum_{t} \log p_\phi(z_t \mid o, z_{<t}).
\label{eq:summarizer}
\end{equation}

\textbf{SAM3 Based Segmentation Integration.}
The SAM3 instruction field is parsed into a target entity list $\{e_1,\ldots,e_K\}$ and used to prompt SAM3 to produce pixel level segmentation over video frames
\begin{equation}
\mathbf{M}_k = \text{SAM3}(\mathbf{v}, e_k), \quad k=1,\ldots,K,
\label{eq:SAM3}
\end{equation}
where $\mathbf{M}_k$ denotes the segmentation mask for the $k$th entity across frames.

\begin{figure}[H]
\centering
\includegraphics[width=\columnwidth]{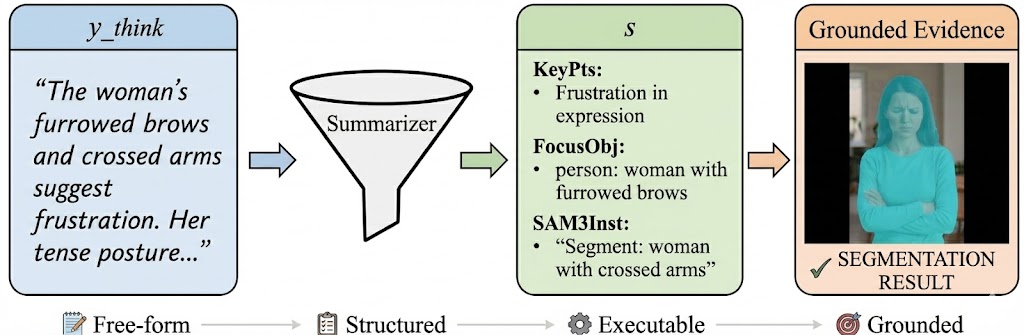}
\caption{Visualization of free-form reasoning transformed into structured evidence, i.e., key points, focus objects, and SAM3 instructions for SAM3 based pixel level grounding.}
\label{fig:summarizer}
\end{figure}

\textbf{Verification Framework.}
Given the SAM3 outputs, we verify whether the reasoning attends to the correct visual regions. For example, if the reasoning states that the man's expression shifts from surprise to relief, then the SAM3 instruction should include the corresponding person entity, and SAM3 should successfully segment the target. The segmentation success rate provides an external indicator of reasoning trustworthiness by establishing an externally auditable linkage between textual evidence claims and pixel level regions across frames. The full reasoning verification pipeline is illustrated in Fig.~\ref{fig:summarizer}.

\section{Experiments}

In this section, we report experimental results and provide further analysis. Implementation details are provided in Appendix.

\subsection{Experimental Setup}
\label{sec:setup}

\textbf{Training Data and Configuration.}
We use Qwen2.5-Omni-7B-Thinker as the base model. To obtain a stable and controllable policy initialization, we follow a standard multimodal reasoning alignment recipe for structured output initialization, enabling the model to consistently produce the three part output $\langle context, think, answer \rangle$. This stage serves to establish a strong baseline and control variables. We apply cold start SFT to stabilize long chain reasoning and format consistency, and then conduct two stage GRPO training to suppress reasoning collapse and format drift. This stage does not include the proposed rewards $R_{\text{ppl}}$ and $R_{\text{tmp}}$. The training corpus contains 24K video audio samples from Video-R1~\cite{feng2025video}, Social-IQ 2.0~\cite{siq2} using the training split, and EMER~\cite{lian2023explainable}. During GRPO training, we sample $G{=}4$ candidate responses per instance, set $\varepsilon{=}0.2$, use a learning rate of $1\times 10^{-6}$, and cap the maximum output length at 2048 tokens. We set reward weights to $\lambda_p{=}0.2$ and $\lambda_t{=}0.2$. All experiments run on \mbox{4$\times$A100 80GB GPUs} with DeepSpeed ZeRO Stage 2 for acceleration and memory optimization.

\textbf{Evaluation Benchmarks.}
We evaluate on three multimodal benchmarks. IntentBench is our curated benchmark for human intent and emotion understanding. It contains 633 videos and 2,689 questions drawn from the Social-IQ 2.0, EMER, and MDPE subsets. Daily-Omni~\cite{zhou2025daily} is a daily scenario audio visual QA benchmark with 684 videos and 1,197 questions covering six task categories. WorldSense~\cite{hong2025worldsense} is a world knowledge audio visual QA benchmark with 1,662 videos and 3,172 questions spanning eight domains.

\textbf{Baselines.}
We compare against two groups of models. The first group includes proprietary models, GPT-4o~\cite{hurst2024gpt}, GPT-o1 (think)~\cite{jaech2024openai}, Gemini-2.5-Pro (think)~\cite{google2025gemini25propreview}, Gemini 2.0 Flash, Gemini 2.0 Flash Lite, Gemini 1.5 Pro~\cite{team2024gemini}, and Claude 3.5 Sonnet~\cite{anthropic2024claude3family}. The second group includes open source omni modal models, Qwen2.5-Omni~\cite{xu2025qwen2}, MiniCPM-o~\cite{yao2024minicpm}, Ola~\cite{liu2025ola}, HumanOmniV2 (7B)~\cite{yang2025humanomniv2}, Unified-IO-2 (8B)~\cite{lu2024unified}, VideoLLaMA2 (7B)~\cite{cheng2024videollama}, and VITA-1.5 (7B)~\cite{fu2025vita}.

\subsection{Main Results}
\label{sec:main_results}

\textbf{IntentBench Results.}
Table~\ref{tab:intentbench} reports results on IntentBench. AffectOmni reaches an average accuracy of 71.89, exceeding the strongest open source baseline HumanOmniV2 (7B) at 69.23. The improvements are moderate but structured, with larger gains on Emotion and How, which are more directly aligned with people-centric evidence. The largest gains occur on Emotion and How, with increases of 4.66 percentage points (pp) and 3.64~pp, the absolute difference between two accuracy percentages. We attribute these gains to the stronger reliance of these categories on people related fine-grained cues and temporally organized reasoning. Compared with prior models that can answer correctly while bypassing human evidence, AffectOmni more reliably forms people-centric evidence chains.

\begin{table*}[h]
\centering
\caption{\textbf{Comparison on IntentBench.} We report category-wise intent understanding
and social reasoning performance. Models marked with $^\dagger$ are proprietary.
The last three rows are ablation variants of AffectOmni built incrementally on
HumanOmniV2 by adding People-Focus ($R_{\text{ppl}}$), Temporal-Order
($R_{\text{tmp}}$), and both rewards jointly (Full Model).}
\scriptsize
\setlength{\tabcolsep}{3.6pt}
\renewcommand{\arraystretch}{0.92}
\resizebox{\linewidth}{!}{
\begin{tabular}{lc|cccccccc|c}
\toprule
\textbf{Methods} & \textbf{LMM} &
\textbf{Why} & \textbf{How} & \textbf{What} & \textbf{When} & \textbf{Who/Which} & \textbf{Other} & \textbf{Emotion} & \textbf{Deception} & \textbf{Avg} \\
\midrule
\multicolumn{11}{c}{\textit{Proprietary MLLMs}} \\
\midrule
GPT-4o$^\dagger$              & -- & 61.46 & 55.69 & 60.00 & 35.71 & 76.00 & 63.31 & 60.99 & 59.00 & 59.98 \\
GPT-o1$^\dagger$ (think)      & -- & 68.19 & 65.82 & 66.04 & 57.14 & 76.00 & 68.83 & 67.26 & 59.50 & 66.69 \\
Gemini-2.5-Pro$^\dagger$ (think) & -- & 68.57 & 67.41 & 65.12 & 57.14 & 64.00 & 70.03 & 68.23 & 60.00 & 67.15 \\
\midrule
\multicolumn{11}{c}{\textit{Open-Source MLLMs}} \\
\midrule
MiniCPM-o                  & 8B & 57.87 & 53.48 & 57.14 & 57.14 & 68.00 & 61.14 & 23.85 & 49.50 & 54.51 \\
VITA-1.5                   & 7B & 53.15 & 49.36 & 51.66 & 71.42 & 64.00 & 61.14 & 53.20 & 59.50 & 54.17 \\
Ola                        & 7B & 60.60 & 55.37 & 56.87 & 64.28 & 76.00 & 62.91 & 46.66 & 44.50 & 57.41 \\
Qwen2.5-Omni               & 7B & 62.60 & 63.44 & 63.53 & 57.14 & 76.00 & 69.03 & 59.74 & 63.50 & 64.20 \\
HumanOmniV2                & 7B & 66.05 & 64.08 & 68.75 & 57.14 & 76.00 & 74.56 & 80.78 & 66.50 & 69.23 \\
\midrule
\multicolumn{11}{c}{\textit{AffectOmni Ablation (built on HumanOmniV2)}} \\
\midrule
\textbf{+ People-Focus($R_{\text{ppl}}$)}  & 7B & 66.48 & 67.25 & 68.33 & 57.14 & 76.00 & 75.35 & 84.23 & 63.50 & 69.78 \\
\textbf{+ Temporal-Order($R_{\text{tmp}}$)} & 7B & 64.33 & 68.20 & 68.96 & 57.14 & 80.00 & 74.16 & 83.67 & 60.50 & 69.62 \\
\rowcolor{gray!12}
\textbf{Full Model (AffectOmni)} & 7B & \textbf{66.91} & \textbf{67.72} & \textbf{70.21} & \textbf{71.43} & 76.00 & 74.95 & \textbf{85.44} & 62.50 & \textbf{71.89} \\
\bottomrule
\end{tabular}
}
\vspace{2pt}
\label{tab:intentbench}
\vspace{-6pt}
\end{table*}

\textbf{Daily-Omni Results.}
Table~\ref{tab:dailyomni} reports results on Daily-Omni across six task dimensions. Compared with the strongest open source baseline HumanOmniV2 (7B), AffectOmni improves average accuracy from 58.47 to 61.90 and yields consistent gains across several representative dimensions. In particular, Context increases by 4.15~pp, Reason increases by 4.01~pp, and 60s tasks increase by 5.46~pp. On Infer, AffectOmni is slightly lower than HumanOmniV2, suggesting that this dimension relies more on general inference and knowledge transfer than on people-centric evidence constraints. We also note that the proprietary model Gemini~2.0~Flash reaches an average accuracy of 67.84 on this benchmark, indicating that performance under complex daily distributions remains influenced by model scale and general capability. Nevertheless, AffectOmni improves over open source 7B scale baselines on several key dimensions, suggesting transfer beyond the diagnostic set, although the gain is not uniform across all dimensions.

\begin{table}[h]
\centering
\caption{Comparison with existing models on Daily-Omni. Models marked
with $\dagger$ are proprietary. We report accuracy (\%) on
representative sub-dimensions; full results are provided in the
supplemental material.}
\label{tab:dailyomni}
\resizebox{\columnwidth}{!}{
\begin{tabular}{lccccc}
\toprule
\textbf{Method} & \textbf{Context} & \textbf{Infer.} & \textbf{Reason.} & \textbf{60s} & \textbf{Avg} \\
\midrule
Gemini 2.0 Flash$^\dagger$      & 63.73 & 76.62          & 75.43          & 56.57          & 67.84 \\
Gemini 2.0 Flash Lite$^\dagger$ & 58.03 & 74.03          & 72.00          & 53.01          & 61.32 \\
\midrule
Unified-IO-2 (8B)   & 26.42 & 35.06          & 29.71          & 30.00          & 28.24 \\
VideoLLaMA2 (7B)    & 35.75 & 40.91          & 34.29          & 31.82          & 35.17 \\
Qwen2.5-Omni (7B)   & 38.86 & 57.79          & 61.71          & 38.36          & 47.45 \\
Ola (7B)            & 39.89 & 61.03          & 66.28          & 48.72          & 49.87 \\
MiniCPM-o (7B)      & 49.22 & 68.83          & 61.14          & 52.00          & 53.13 \\
HumanOmniV2 (7B)    & 51.81 & \textbf{72.72} & 74.28          & 53.09          & 58.47 \\

\midrule

+ People-Focus ($R_\mathrm{ppl}$)    
                    & 58.55 & 75.32          & 77.14          & 60.36          & 62.57 \\
+ Temporal-Order ($R_\mathrm{tmp}$)  
                    & 55.96 & 73.38          & 77.71          & 58.73          & 62.41 \\
\rowcolor{gray!12}
\textbf{AffectOmni (Ours)} 
                    & \textbf{55.96} & 72.08 & \textbf{78.29} & \textbf{58.55} & \textbf{61.90} \\
\bottomrule
\end{tabular}
}
\end{table}

\textbf{WorldSense Results.}
Table~\ref{tab:worldsense} reports domain wise performance on WorldSense. AffectOmni achieves an average accuracy of 48.80, exceeding the open source baseline HumanOmniV2 (7B) at 47.70 and slightly surpassing the proprietary baseline Gemini1.5 Pro. The largest domain wise gain appears in Music, with an improvement of 3.9~pp. Compared with Gemini1.5 Pro, AffectOmni remains behind on Film, Tech, and Perform., which rely more on external knowledge and cross domain semantics. At the same time, AffectOmni shows a clear advantage in Music, leading to a higher overall average. These results suggest that our gains primarily arise from improved reasoning structures that organize people related cues and temporal evidence, rather than from uniform improvements across knowledge intensive domains.

\begin{table}[h]
\centering
\caption{Comparison with existing models on WorldSense. Models marked
with $\dagger$ are proprietary. We report accuracy (\%) on
representative domains; full results are provided in the supplemental
material.}
\label{tab:worldsense}
\resizebox{\columnwidth}{!}{
\begin{tabular}{lccccc}
\toprule
\textbf{Method} & \textbf{Tech} & \textbf{Film} & \textbf{Perform.} & \textbf{Music} & \textbf{Avg} \\
\midrule
Claude 3.5 Sonnet$^\dagger$ & 43.70 & 36.50 & 30.70 & 33.90 & 34.80 \\
GPT-4o$^\dagger$            & 48.00 & 43.50 & 41.90 & 42.70 & 42.60 \\
Gemini 1.5 Pro$^\dagger$    & 53.70 & 50.40 & 52.40 & 42.00 & 48.00 \\
\midrule
Unified-IO-2 XXL (7B) & 27.10 & 23.70 & 25.50 & 27.30 & 25.90 \\
VideoLLaMA2 (7B)      & 29.40 & 24.50 & 26.20 & 27.10 & 25.40 \\
VITA-1.5 (7B)         & 38.20 & 39.80 & 41.20 & 39.90 & 36.90 \\
Qwen2.5-Omni (7B)     & 47.80 & 43.80 & 48.30 & \textbf{47.30} & 45.40 \\
HumanOmniV2 (7B)      & 49.60 & 47.50 & 48.40 & 43.30 & 47.70 \\

\midrule
+ People-Focus ($R_\mathrm{ppl}$)   
                      & 50.90 & \textbf{50.40} & 46.50 & 46.70 & 47.60 \\
+ Temporal-Order ($R_\mathrm{tmp}$) 
                      & 51.30 & 48.50 & 48.80 & 46.40 & 48.60 \\
\rowcolor{gray!12}
\textbf{AffectOmni (Ours)} 
                      & \textbf{51.50} & 49.30 & \textbf{50.40} & 47.20 & \textbf{48.80} \\
\bottomrule
\end{tabular}
}
\vspace{-8pt}
\end{table}

\subsection{Diagnostic Analysis}
\label{sec:diagnostic}

\textbf{Effectiveness of Reward Components.}
To assess how the People-Focus reward $R_{\text{ppl}}$ and the Temporal-Order reward $R_{\text{tmp}}$ mitigate shortcut reasoning and improve verifiability, we conduct an ablation on IntentBench that varies only reward configurations. Table~\ref{tab:reward_ablation} shows that the Baseline with accuracy-only reward reaches 69.23 in Avg. Adding $R_{\text{ppl}}$ increases Avg. to 69.78 and improves Emotion to 84.23, indicating that the People-Focus constraint more reliably encourages reliance on people-related cues. Adding $R_{\text{tmp}}$ alone yields an Avg. of 69.62 and improves Emotion to 83.67, suggesting that explicit temporal constraints help organize reasoning over time-sensitive evidence.

Finally, the full model reaches 71.89 in Avg., improving by 2.66~pp over the baseline, and increases the When category by 14.29~pp, which confirms the importance of the complete reward design for temporally sensitive questions.  We also observe a decrease on Deception. Since the bootstrap confidence interval includes zero, we do not interpret it as a statistically reliable degradation. Instead, it suggests a boundary of the current inductive bias, as deception and sarcasm often require modeling the mismatch between surface behavior and underlying intent. Lastly, although the Baseline, $R_{\text{ppl}}$, and $R_{\text{tmp}}$ all obtain 57.14 on When, this does not imply saturation. These settings likely fall into the same granularity bin, whereas the full model moves beyond multiple granularity levels, further suggesting that joint rewards substantially improve temporal reasoning.

\begin{table}[h]
\centering
\caption{We report accuracy (\%) on IntentBench across key categories.
All variants share the same base model and training setup, and differ only in the enabled reward terms, showing the marginal gain brought by each component.}
\label{tab:reward_ablation}
\resizebox{\linewidth}{!}{
\begin{tabular}{lccccc}
\toprule
\textbf{Configuration} & \textbf{Emotion} & \textbf{Deception} & \textbf{When} & \textbf{Why} & \textbf{Avg} \\
\midrule
Baseline (Acc-only) & 80.78 & 66.50 & 57.14 & 66.05 & 69.23 \\
+ People-Focus ($R_{\text{ppl}}$) & 84.23 & 63.50 & 57.14 & 66.48 & 69.78 \\
+ Temporal-Order ($R_{\text{tmp}}$) & 83.67 & 60.50 & 57.14 & 64.33 & 69.62 \\
\rowcolor{gray!15}
\textbf{Full Model} & \textbf{85.44} & \textbf{62.50} & \textbf{71.43} & \textbf{66.91} & \textbf{71.89} \\
\bottomrule
\end{tabular}
}
\vspace{-8pt}
\end{table}

\textbf{Comparative Scoring Analysis.}
We find that when the reward distribution becomes overly concentrated due to score clustering, or drifts across samples due to calibration drift, the advantage gaps between candidates are compressed. This reduces the effectiveness of GRPO~\cite{shao2024deepseekmath} updates. To address this issue, we compare independent absolute scoring with comparative scoring that enforces ranking, focusing on reward discriminability and downstream performance. Because the two schemes use different raw score scales, we apply min max normalization to each set of scores and linearly rescale them to the common interval $[1,10]$. This enables fair comparison of CV, defined as Std over Mean, within-group CV, and the shape of the score distribution.

\begin{figure}[h]
\centering
\includegraphics[width=\columnwidth]{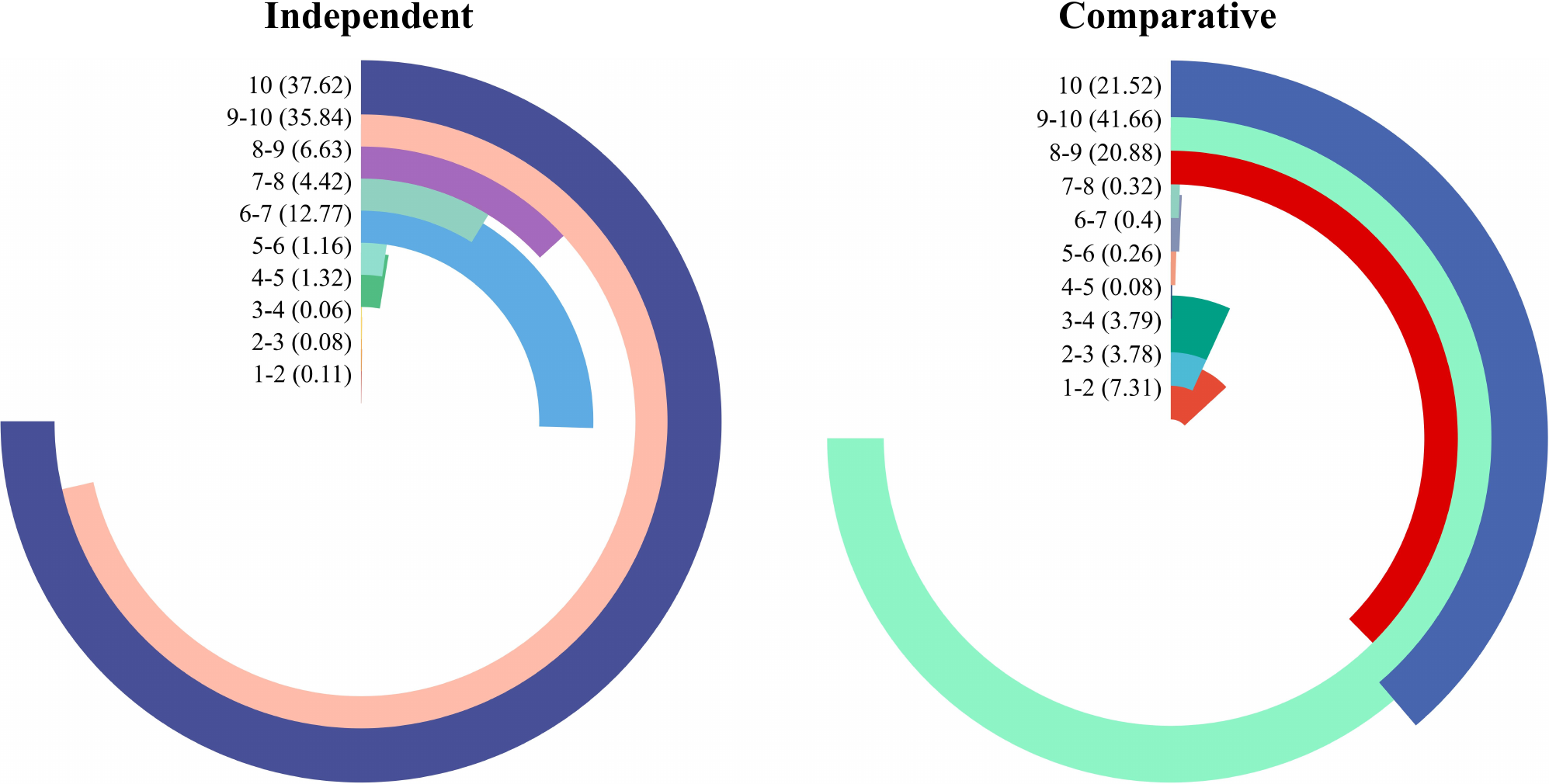}
\caption{\textbf{Reward Signal Discrimination Analysis.} 
Each arc represents a score interval (1--10 scale), with arc length proportional 
to the percentage of samples falling in that interval. 
Left (Independent): LLM as a Judge scores each candidate in isolation, causing 
severe high score clustering, over 73\% of scores fall in the 9--10 range. 
Right (Comparative): within-group comparative scoring forces relative ranking 
among candidates, yielding a more balanced distribution with meaningful 
differentiation across the full score spectrum.}
\label{fig:score_distribution}
\end{figure}

Table~\ref{tab:scoring_comparison} shows that comparative scoring substantially increases dispersion and within-group separability of reward signals relative to independent scoring. The overall CV increases from 0.153 to 0.325, and the within-group CV increases from 0.049 to 0.138. These results indicate clearer relative differences among candidates within the same group, which yields a more informative and stable update signal for policy optimization. Consistent with these statistics, the unified scale visualization in Fig.~\ref{fig:score_distribution} reveals pronounced clustering in the high score range under independent scoring. In contrast, comparative scoring reduces perfect score clustering, allocates more probability mass to the 9 to 10 and 8 to 9 ranges, and maintains non zero coverage in the low score tail. For example, the 1 to 2 bin has 7.31. This makes extreme calibration shifts caused by absolute scale drift less likely. Correspondingly, IntentBench average accuracy improves from 69.78 to 71.89, showing that stronger reward discriminability translates into downstream reasoning gains.

\begin{table}[h]
\centering
\caption{Comparison of scoring strategies. ``Ind.'' denotes independent scoring. ``Comp.'' denotes our comparative scoring. CV is the coefficient of variation, computed as standard deviation over mean.}
\label{tab:scoring_comparison}
\begin{tabular}{@{}lcc@{}}
\toprule
\textbf{Metric} & \textbf{Independent} & \textbf{Comparative} \\
\midrule
CV & 0.153 & 0.325 \\
Within-Group CV & 0.049 & 0.138 \\
\midrule
IntentBench Avg (\%) & 69.78 & 71.89 \\
\bottomrule
\end{tabular}
\end{table}

\begin{figure*}[t]
\centering
\includegraphics[width=\textwidth,height=0.4\textheight]{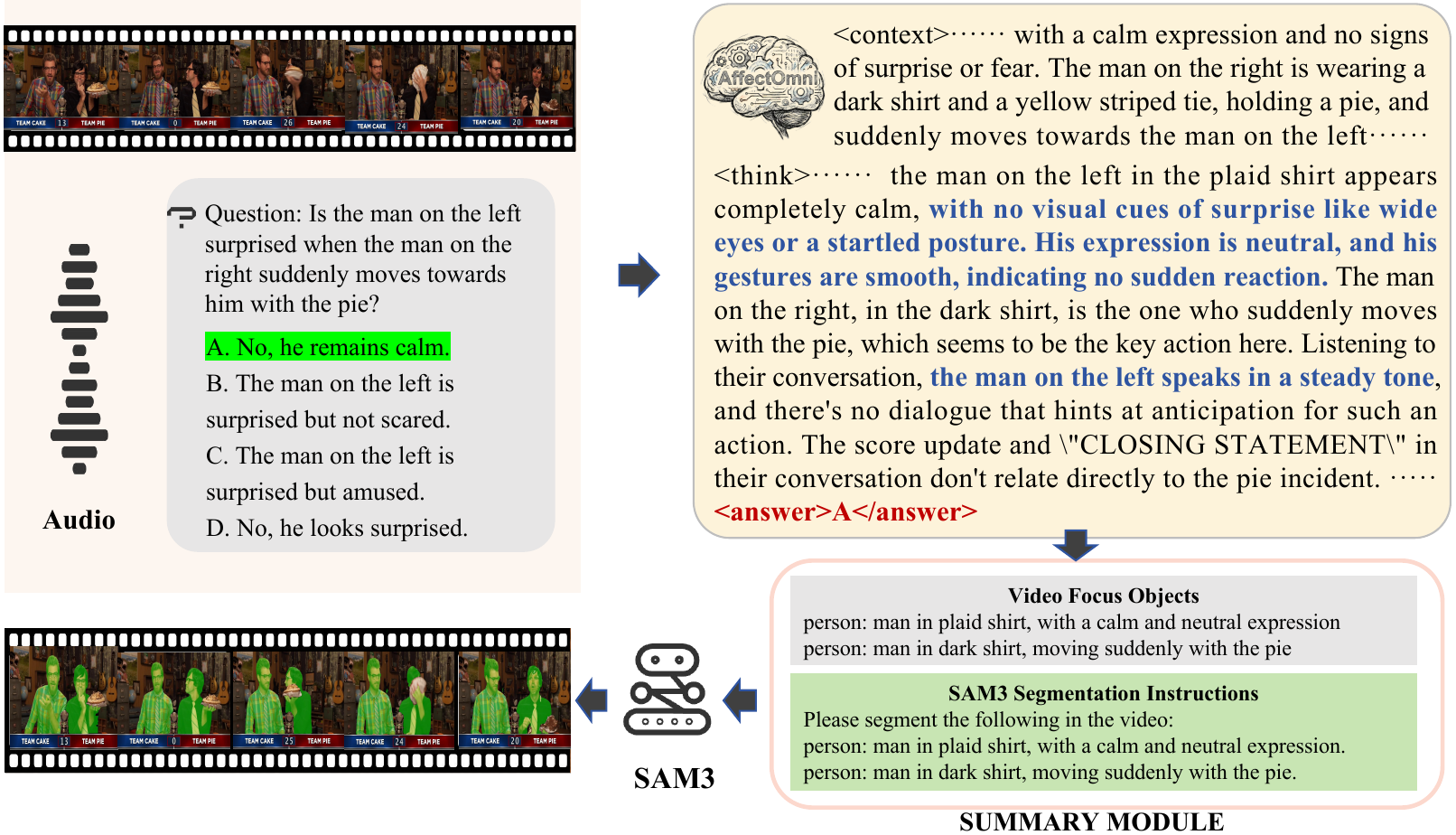}
\caption{\textbf{Reasoning to evidence grounding framework.} The framework connects structured reasoning chains to pixel-level grounding via SAM3: AffectOmni summarizes focus entities from the reasoning, converts them into segmentation instructions, and visualizes the exact regions to enable automatic consistency checks and human inspection.}
\label{fig:closed_loop}
\vspace{-5.5mm}
\end{figure*}

\textbf{Reasoning Verification Framework.}
A core challenge in affective reasoning is to verify whether a model truly attends to the visual evidence it cites in its analysis. To address this challenge, we introduce a reasoning verification framework in AffectOmni that connects reasoning with visual localization. As shown in Fig.~\ref{fig:closed_loop}, given a video and the question ``Is the man on the left surprised?'', AffectOmni generates a structured reasoning chain. It identifies that the man in a checkered shirt appears calm, with no visual cues of surprise such as widened eyes or a startle posture. The reasoning is then passed to a summarization module that extracts the focus entities in the video, for example the man in a checkered shirt with a calm and neutral expression, and converts them into SAM3 segmentation instructions. Finally, SAM3 produces pixel level masks that visualize the regions referenced by the reasoning. This provides a verifiable linkage between the language description and visual evidence in the video frames, enabling automatic consistency checks and human inspection of whether the claimed content is grounded in the attended regions.

\textbf{Statistical Significance of Performance Gains.}
To verify that the observed gains reflect systematic effects of the
proposed reward design rather than random variation, this study
conducts one-sided paired bootstrap tests with 10{,}000 resamples
on the full IntentBench test set ($n = 2{,}689$), with results
reported in Table~\ref{tab:bootstrap}.

AffectOmni achieves statistically significant improvements in both
categories directly targeted by $R_\mathrm{ppl}$. The Emotion
category yields the largest absolute gain at $+4.66$~pp ($p =
0.008$, 95\% CI $[+1.2,\ +8.3]$~pp), while How achieves the
strongest significance at $p = 0.002$ with $+3.64$~pp ($[+1.3,\
+5.9]$~pp). The lower $p$-value on How, despite its smaller
absolute gain, is explained by its much larger sample size ($n =
632$ vs.\ $n = 133$), which provides greater statistical power.
Both confidence intervals exclude zero, confirming that these gains
arise from the reward mechanism rather than sampling noise.

Beyond the directly targeted categories, Why, What, and Other show
positive but non-significant trends ($+0.86$, $+1.46$, $+0.39$~pp)
with confidence intervals spanning zero, indicating partial but
unreliable transfer of people-centric reasoning capabilities. The
Deception category stands apart with a decline of $-4.00$~pp ($p =
1.000$, 95\% CI $[-9.0,\ +1.0]$~pp), the wide interval reflects
high within-category variance, so the drop cannot be distinguished
from random variation. Deception requires detecting mismatches
between surface behavior and underlying intent, a pragmatic
reasoning demand where stronger reliance on observable affective
cues may reduce rather than increase sensitivity.

\begin{table}[t]
  \caption{One-sided paired bootstrap significance test (10{,}000
  resamples) on IntentBench ($n=2{,}689$), testing whether
  AffectOmni $>$ HumanOmniV2 per category.
  $\Delta$ denotes the mean accuracy difference
  (AffectOmni $-$ HumanOmniV2).
  Shaded rows are categories directly targeted by $R_\mathrm{ppl}$.
  $^{**}p{<}0.01$;\ $^{*}p{<}0.05$.}
  \label{tab:bootstrap}
  \centering
  \begin{tabular}{lrccc}
    \toprule
    Category & $n$ & $\Delta$ (pp) & 95\% CI (pp) & $p$ \\
    \midrule
    \rowcolor{gray!15}
    Emotion ($\leftarrow R_\mathrm{ppl}$) & 133 & $+4.66^{**}$ & $[+1.2,\ +8.3]$ & $0.008$ \\
    \rowcolor{gray!15}
    How ($\leftarrow R_\mathrm{ppl}$)     & 632 & $+3.64^{**}$ & $[+1.3,\ +5.9]$ & $0.002$ \\
    \midrule
    Why                                    & 698 & $+0.86$      & $[-1.4,\ +3.2]$ & $0.484$ \\
    What                                   & 480 & $+1.46$      & $[-1.2,\ +4.2]$ & $0.317$ \\
    Other                                  & 507 & $+0.39$      & $[-1.8,\ +2.6]$ & $0.779$ \\
    Deception                              & 200 & $-4.00$      & $[-9.0,\ +1.0]$ & $1.000$ \\
    \bottomrule
  \end{tabular}
\vspace{-8pt}
\end{table}

\textbf{Cross-Domain Generalization on General Video QA.}
A natural concern is whether affective RL training causes catastrophic
forgetting of general video understanding capabilities. To test this,
this study evaluates AffectOmni on NExT-QA~\cite{xiao2021next}, a
general-purpose video QA benchmark covering causal, temporal, and
descriptive reasoning, without any domain-specific fine-tuning.

As shown in Table~\ref{tab:nextqa_overall}, AffectOmni achieves
80.52\% overall accuracy on the multiple-choice test set, surpassing
HumanOmniV2 (79.78\%) and several general-purpose video models of
comparable scale. Fine-grained results in Table~\ref{tab:nextqa_detail}
further reveal that the improvements are not uniform. AffectOmni
outperforms HumanOmniV2 on the Temporal subset (79.44\% vs.\ 76.50\%)
and the Descriptive subset (86.48\% vs.\ 83.77\%), while HumanOmniV2
retains a slight advantage on Causal (80.50\% vs.\ 79.28\%). The
gains on Temporal and Descriptive are consistent with the design
intent of this study: People-Focus and Temporal-Order rewards train
transferable capabilities in entity anchoring, temporal organization,
and evidence chain construction, which benefit reasoning tasks beyond
affective inference. The modest Causal deficit suggests that causal
inference relies more on general semantic priors not directly
addressed by the reward design. Together, these results indicate
that people-centric RL training does not degrade general video
understanding and instead produces structured reasoning capabilities
that transfer positively across task domains. Representative qualitative comparisons in Appendix~F further illustrate the
improvements in micro-expression analysis and temporal affective reasoning.

\begin{table}[t]
  \caption{Results on NExT-QA multiple-choice test set. AffectOmni
  surpasses HumanOmniV2 without domain-specific fine-tuning,
  confirming that affective RL training does not cause catastrophic
  forgetting of general video understanding.}
  \label{tab:nextqa_overall}
  \centering
  \begin{tabular}{lcc}
    \toprule
    Method & Size & Overall Acc.\ (\%) \\
    \midrule
    LLaVA-Video~\cite{zhang2024llava}          & 7B & 83.20 \\
    PLLaVA~\cite{xu2024pllava}                 & 7B & 81.00 \\
    Magma~\cite{yang2025magma}                 & 8B & 80.90 \\
    \textbf{AffectOmni (Ours)}                 & \textbf{7B} & \textbf{80.52} \\
    HumanOmniV2~\cite{yang2025humanomniv2}     & 7B & 79.78 \\
    LLaVA-OneVision~\cite{li2024llava}         & 7B & 79.40 \\
    mPLUG-Owl3~\cite{ye2024mplug}              & 8B & 78.60 \\
    VideoChat2~\cite{li2024mvbench}            & 8B & 63.20 \\
    \bottomrule
  \end{tabular}
\vspace{-6pt}
\end{table}

\begin{table}[t]
  \caption{Fine-grained NExT-QA results by question type. AffectOmni
  shows gains over HumanOmniV2 on Temporal and Descriptive subsets,
  consistent with the transferable reasoning capabilities trained by
  People-Focus and Temporal-Order rewards.}
  \label{tab:nextqa_detail}
  \centering
  \begin{tabular}{lccc}
    \toprule
    Method & Causal & Temporal & Descriptive \\
    \midrule
    AffectOmni  & 79.28 & \textbf{79.44} & \textbf{86.48} \\
    HumanOmniV2 & \textbf{80.50} & 76.50 & 83.77 \\
    \bottomrule
  \end{tabular}
\vspace{-10pt}
\end{table}

\section{Discussion}
\label{sec:discussion}
This section evaluates the Minimal Evidence Package (MEP) as an Executable Evidence Interface (EEI). We emphasize that its goal is not segmentation quality itself, but converting reasoning into an executable and checkable evidence carrier for post-hoc auditing. This abstraction is important in high stakes scenarios such as clinical decision support or affective companion systems. In such settings, a system should output not only conclusions but also evidence packages that are auditable and reviewable. This supports explanation and provides a safety fallback.

MEP encourages reasoning to produce locatable elements, for example the focus person, key actions, and temporal cues, and maps them into executable verification instructions. MEP interfaces with downstream verifiers through the EEI. The interface layer uses \{FocusObj, TimeSpan, EvidenceType\} as a minimal query form, and heterogeneous verifiers can be integrated with lightweight adaptation. This enables reuse of the auditing interface without modifying the upstream reasoning compression core. Our current implementation uses SAM3 pixel level localization as one instance. The same interface can extend to keyframe retrieval, object tracking, speaker segment localization, and textual evidence retrieval.

To validate these claims and quantify MEP as an EEI rather than a display component, we design three lightweight diagnostic experiments without retraining. These experiments probe evidence extractability, evidence to verification consistency, and falsifiability under external auditing.

\subsection{Evidence Extractability}

The first experiment tests whether the reasoning chain provides a stable basis of structured information. We randomly sample 100 think outputs from IntentBench and tally three evidence elements that are directly relevant to generating verification instructions. Visual evidence refers to people related observable cues such as expression, gaze, and posture. Temporal evidence refers to order, change, or stage descriptions over time. Behavioral evidence refers to executable actions or interaction cues such as gestures, approach and avoidance, and conversational interaction. To avoid missed detections and ambiguity from keyword matching, we use Qwen-Max as the judge and apply a strict binary criterion. We also require a minimal supporting span and apply schema constrained parsing to restrict the output format and reduce drift from free form generation. This statistic is used to diagnose interface feasibility rather than as a final evaluation metric.

Table~\ref{tab:evidence_extractability} and Table~\ref{tab:evidence_by_type} show that the occurrence rates of the three evidence types are 84\%, 72\%, and 89\%, and that 93\% of samples contain at least two evidence types. These results indicate that the model produces composable evidence primitives for most samples, which provides the information basis for the Summarizer to compress reasoning into an MEP.

\begin{table}[h]
\centering
\small
\setlength{\tabcolsep}{5pt}
\caption{Occurrence rates of three types of structured evidence (Visual/Temporal/Behavioral) and coverage rate of $\geq$2 evidence types from 100 randomly sampled \texttt{think} outputs.}
\label{tab:evidence_extractability}
\begin{tabular}{lcc}
\toprule
\textbf{Evidence Type} & \textbf{Count} & \textbf{Rate} \\
\midrule
Visual Evidence & 84 & 84\% \\
Temporal Evidence & 72 & 72\% \\
Behavioral Evidence & 89 & 89\% \\
\midrule
$\geq$2 Types & 93 & 93\% \\
\bottomrule
\end{tabular}
\vspace{-8pt}
\end{table}

\begin{table}[h]
\centering
\small
\setlength{\tabcolsep}{4pt}
\caption{Evidence occurrence rates grouped by question type.}
\label{tab:evidence_by_type}
\begin{tabular}{lcccc}
\toprule
\textbf{Type} & \textbf{Samples} & \textbf{Visual} & \textbf{Temporal} & \textbf{Behavioral} \\
\midrule
Other & 28 & 75\% & 75\% & 89\% \\
Why & 27 & 81\% & 70\% & 81\% \\
How & 19 & 95\% & 74\% & 89\% \\
What & 14 & 86\% & 64\% & 93\% \\
Deception & 8 & 88\% & 62\% & 100\% \\
Emotion & 4 & 100\% & 100\% & 100\% \\
\midrule
\rowcolor{gray!15}
\textbf{Overall} & \textbf{100} & \textbf{84\%} & \textbf{72\%} & \textbf{89\%} \\
\bottomrule
\end{tabular}
\vspace{-8pt}
\end{table}

\subsection{Evidence to Verification Consistency}

The second experiment evaluates the operationality of the EEI, namely whether the focus object description in think provides sufficient detail for successful localization by downstream visual verifiers. We count elements including person mentions, appearance attributes such as color and clothing, visual features such as expression and posture, and locatable actions. Based on these elements, we define two grounding criteria. Basic Grounding requires an explicitly mentioned and segmentable person. High Quality Grounding requires a person mention and at least one executable localization cue, either a visual feature or a locatable action.

Table~\ref{tab:consistency} shows that 98\% of samples mention a person, 91\% include visual feature descriptions, and 94\% include locatable actions. Accordingly, 98\% of samples satisfy Basic Grounding and 86\% satisfy High Quality Grounding. By question type in Table~\ref{tab:grounding_by_type}, Basic Grounding ranges from 93\% to 100\% across categories, and High Quality Grounding ranges from 82\% to 100\%, which indicates consistent executability across question types. Appearance attributes appear in only 39\% of cases, which is markedly lower than visual features and action cues. This suggests that current reasoning relies more on dynamic evidence such as expression, posture, and actions. Under occlusion or in crowded multi person scenes, grounding can become more ambiguous without additional discriminative appearance cues or spatial references.

\begin{table}[h]
\centering
\small
\setlength{\tabcolsep}{4pt}
\caption{Occurrence rates of grounding-relevant elements and two levels of grounding criteria from 100 randomly sampled reasoning chains.}
\label{tab:consistency}
\begin{tabular}{lcc}
\toprule
\textbf{Indicator} & \textbf{Count} & \textbf{Rate} \\
\midrule
Mentions Person & 98 & 98\% \\
Has Appearance (color/clothing) & 39 & 39\% \\
Has Visual Feature (expression/posture) & 91 & 91\% \\
Has Locatable Action & 94 & 94\% \\
\midrule
\rowcolor{gray!15}
\textbf{Basic Grounding} & \textbf{98} & \textbf{98\%} \\
\rowcolor{gray!15}
\textbf{High-Quality Grounding} & \textbf{86} & \textbf{86\%} \\
\bottomrule
\end{tabular}
\vspace{-8pt}
\end{table}

\begin{table}[h]
\centering
\small
\setlength{\tabcolsep}{4pt}
\caption{Grounding attainment rates grouped by question type.}
\label{tab:grounding_by_type}
\begin{tabular}{lccc}
\toprule
\textbf{Type} & \textbf{Samples} & \textbf{Basic} & \textbf{High-Quality} \\
\midrule
Other & 28 & 96\% & 82\% \\
Why & 27 & 100\% & 85\% \\
How & 19 & 100\% & 89\% \\
What & 14 & 93\% & 86\% \\
Deception & 8 & 100\% & 88\% \\
Emotion & 4 & 100\% & 100\% \\
\midrule
\rowcolor{gray!15}
\textbf{Overall} & \textbf{100} & \textbf{98\%} & \textbf{86\%} \\
\bottomrule
\end{tabular}
\vspace{-8pt}
\end{table}

\begin{table}[b]
\centering
\caption{Annotators judge, based only on SAM3 segmentation, whether it is related to the question target, and we report the contingency table between this relevance judgment and answer correctness, together with conditional accuracy (n=100).}
\label{tab:counterfactual}
\resizebox{0.9\columnwidth}{!}{
\begin{tabular}{l c c c}
\toprule
& \textbf{Seg. Relevant} & \textbf{Seg. Irrelevant} & \textbf{Total} \\
\midrule
Answer Correct & 55 & 12 & 67 \\
Answer Wrong & 11 & 22 & 33 \\
\midrule
Conditional Acc. & 83.3\% & 35.3\% & 67\% \\
\bottomrule
\end{tabular}
}
\vspace{1pt}
\end{table}

\subsection{External Auditing and Falsifiability}

The third experiment evaluates falsifiability through external auditing. It tests whether the auditing interface provides an external signal that helps
assess answer trustworthiness, rather than producing superficially plausible
visualizations for arbitrary instructions. We randomly sample 100 items from IntentBench and ask annotators to view only the question and the SAM3 segmentation results. Annotators do not have access to the model textual reasoning or the ground truth label. They provide a binary relevance judgment, relevant or irrelevant, on whether the segmentation focuses on objects related to the question target. Relevant means that the segmented subject matches the asked for object and covers the key person.

Table~\ref{tab:counterfactual} shows that when segmentation is judged relevant, conditional accuracy is 83.3\%. When segmentation is judged irrelevant, conditional accuracy drops to 35.3\%, a difference of 48.0 percentage points. The contingency table yields an overall agreement rate of 77\%, computed as $(55+22)/100$, and the Phi coefficient is $\phi=0.48$, which indicates a moderate positive association between focusing on the key object and answering correctly. Among the 67 correct-answer cases, 12 have irrelevant segmentation, corresponding to a 17.9\% false-negative rate. These false negatives reduce diagnostic coverage but do not affect training or final answer accuracy because the verifier is post-hoc and excluded from GRPO optimization. Such errors may come from summarization bias in evidence-query extraction, SAM3 grounding failure, or downstream relevance-judgment noise. Together, these results show that annotators do not need to read the reasoning text. By inspecting the post-hoc visual evidence alone, they can obtain an external signal for answer reliability, which supports the value of the MEP as an evidence carrier that is external, operational, and auditable. When segmentation decouples from the question target, accuracy declines substantially. This demonstrates falsifiability because evidence mismatch exposes potentially unreliable outputs in an observable manner, rather than always producing seemingly reasonable results. Representative failure cases in Appendix~G further illustrate pragmatic
reasoning failures in sarcasm interpretation and reasoning-selection
disconnects.

\section{Conclusion}
This paper proposes AffectOmni for verifiable affective
reasoning, addressing the trustworthiness bottleneck in
multimodal emotion and intent understanding where models
produce correct answers while bypassing fine grained
people-centric evidence.
We introduce People-Focus and Temporal-Order rewards into
GRPO training to incorporate people-centered evidence selection
and temporal organization into the reinforcement learning
objective, and propose within-group comparative scoring to
mitigate score clustering and calibration drift in
LLM as a Judge evaluation.
A post-hoc reasoning-to-evidence auditing interface further compresses chain-of-thought outputs into executable evidence instructions grounded via SAM3, supporting external consistency and falsifiability diagnostics (Section~\ref{sec:discussion}).
Experiments on IntentBench, Daily-Omni, and WorldSense confirm
consistent gains over open-source 7B scale baselines on
people-centric and temporally sensitive tasks.
Temporal interval localization remains coarse and error
propagation in the verification module is unquantified, future
work will address localization precision, auxiliary verification
rewards, and reward generalizability.

\section*{Acknowledgments}
This work was supported in part by the National Natural Science
Foundation of China (Grant No.~62227807 and Grant No.~U24B20186),
in part by the Brain Science and Brain-like Intelligence
Technology---National Science and Technology Major Project
(No.~2021ZD0200600, No.~2021ZD0200408),
and in part by the WQ \& UCAS Research Academy Intelligent
Computing Center (WRA-ICC) and the Supercomputing Center of
Lanzhou University.

{\appendix[Supplementary Material]
The supplementary material provides additional details and analyses
supporting the main text. Appendix~A describes the full three-stage
training pipeline and Thinking Summarizer configuration.
Appendix~B reports complete per-subcategory results on Daily-Omni
and WorldSense. Appendix~C analyzes how the group size $G$
affects reward discriminability and verifies the absence of
positional bias. Appendix~D examines sensitivity to the reward
weights $\lambda_p$ and $\lambda_t$. Appendix~E evaluates
reasoning chain quality along five diagnostic dimensions using
both automatic and human assessment. Appendices~F and~G present
qualitative comparisons of success and failure cases,
respectively, illustrating the behavioral patterns induced by
the proposed rewards.
}



 
%

\bibliographystyle{IEEEtran}
\bibliography{affectomni_refs}

@article{carion2025sam,
  title={Sam 3: Segment anything with concepts},
  author={Carion, Nicolas and Gustafson, Laura and Hu, Yuan-Ting and Debnath, Shoubhik and Hu, Ronghang and Suris, Didac and Ryali, Chaitanya and Alwala, Kalyan Vasudev and Khedr, Haitham and Huang, Andrew and others},
  journal={arXiv preprint arXiv:2511.16719},
  year={2025}
}

@article{achiam2023gpt,
  title={Gpt-4 technical report},
  author={Achiam, Josh and Adler, Steven and Agarwal, Sandhini and Ahmad, Lama and Akkaya, Ilge and Aleman, Florencia Leoni and Almeida, Diogo and Altenschmidt, Janko and Altman, Sam and Anadkat, Shyamal and others},
  journal={arXiv preprint arXiv:2303.08774},
  year={2023}
}

@article{team2023gemini,
  title={Gemini: a family of highly capable multimodal models},
  author={Team, Gemini and Anil, Rohan and Borgeaud, Sebastian and Alayrac, Jean-Baptiste and Yu, Jiahui and Soricut, Radu and Schalkwyk, Johan and Dai, Andrew M and Hauth, Anja and Millican, Katie and others},
  journal={arXiv preprint arXiv:2312.11805},
  year={2023}
}

@article{xu2025qwen2,
  title={Qwen2. 5-omni technical report},
  author={Xu, Jin and Guo, Zhifang and He, Jinzheng and Hu, Hangrui and He, Ting and Bai, Shuai and Chen, Keqin and Wang, Jialin and Fan, Yang and Dang, Kai and others},
  journal={arXiv preprint arXiv:2503.20215},
  year={2025}
}

@article{huang2025vision,
  title={Vision-r1: Incentivizing reasoning capability in multimodal large language models},
  author={Huang, Wenxuan and Jia, Bohan and Zhai, Zijie and Cao, Shaosheng and Ye, Zheyu and Zhao, Fei and Xu, Zhe and Hu, Yao and Lin, Shaohui},
  journal={arXiv preprint arXiv:2503.06749},
  year={2025}
}

@article{yuan2025vl,
  title={Vl-cogito: Progressive curriculum reinforcement learning for advanced multimodal reasoning},
  author={Yuan, Ruifeng and Xiao, Chenghao and Leng, Sicong and Wang, Jianyu and Li, Long and Xu, Weiwen and Chan, Hou Pong and Zhao, Deli and Xu, Tingyang and Wei, Zhongyu and others},
  journal={arXiv preprint arXiv:2507.22607},
  year={2025}
}

@article{xia2025visionary,
  title={Visionary-r1: Mitigating shortcuts in visual reasoning with reinforcement learning},
  author={Xia, Jiaer and Zang, Yuhang and Gao, Peng and Li, Sharon and Zhou, Kaiyang},
  journal={arXiv preprint arXiv:2505.14677},
  year={2025}
}

@article{yang2025humanomniv2,
  title={HumanOmniV2: From Understanding to Omni-Modal Reasoning with Context},
  author={Yang, Qize and Yao, Shimin and Chen, Weixuan and Fu, Shenghao and Bai, Detao and Zhao, Jiaxing and Sun, Boyuan and Yin, Bowen and Wei, Xihan and Zhou, Jingren},
  journal={arXiv preprint arXiv:2506.21277},
  year={2025}
}

@inproceedings{yu2018mattnet,
  title={Mattnet: Modular attention network for referring expression comprehension},
  author={Yu, Licheng and Lin, Zhe and Shen, Xiaohui and Yang, Jimei and Lu, Xin and Bansal, Mohit and Berg, Tamara L},
  booktitle={Proceedings of the IEEE conference on computer vision and pattern recognition},
  pages={1307--1315},
  year={2018}
}

@inproceedings{reich2023measuring,
    title = "Measuring Faithful and Plausible Visual Grounding in {VQA}",
    author = "Reich, Daniel  and
      Putze, Felix  and
      Schultz, Tanja",
    editor = "Bouamor, Houda  and
      Pino, Juan  and
      Bali, Kalika",
    booktitle = "Findings of the Association for Computational Linguistics: EMNLP 2023",
    month = dec,
    year = "2023",
    address = "Singapore",
    publisher = "Association for Computational Linguistics",
    url = "https://aclanthology.org/2023.findings-emnlp.206/",
    doi = "10.18653/v1/2023.findings-emnlp.206",
    pages = "3129--3144"
}

@article{das2017human,
  title={Human attention in visual question answering: Do humans and deep networks look at the same regions?},
  author={Das, Abhishek and Agrawal, Harsh and Zitnick, Larry and Parikh, Devi and Batra, Dhruv},
  journal={Computer Vision and Image Understanding},
  volume={163},
  pages={90--100},
  year={2017},
  publisher={Elsevier}
}

@inproceedings{favero2024multi,
  title={Multi-modal hallucination control by visual information grounding},
  author={Favero, Alessandro and Zancato, Luca and Trager, Matthew and Choudhary, Siddharth and Perera, Pramuditha and Achille, Alessandro and Swaminathan, Ashwin and Soatto, Stefano},
  booktitle={Proceedings of the IEEE/CVF Conference on Computer Vision and Pattern Recognition},
  pages={14303--14312},
  year={2024}
}

@article{suryavardan2023factify,
  title={Factify 2: A multimodal fake news and satire news dataset},
  author={Suryavardan, S and Mishra, Shreyash and Patwa, Parth and Chakraborty, Megha and Rani, Anku and Reganti, Aishwarya and Chadha, Aman and Das, Amitava and Sheth, Amit and Chinnakotla, Manoj and others},
  journal={arXiv preprint arXiv:2304.03897},
  year={2023}
}

@article{ravi2024sam,
  title={Sam 2: Segment anything in images and videos},
  author={Ravi, Nikhila and Gabeur, Valentin and Hu, Yuan-Ting and Hu, Ronghang and Ryali, Chaitanya and Ma, Tengyu and Khedr, Haitham and R{\"a}dle, Roman and Rolland, Chloe and Gustafson, Laura and others},
  journal={arXiv preprint arXiv:2408.00714},
  year={2024}
}

@article{zhou2025daily,
  title={Daily-Omni: Towards Audio-Visual Reasoning with Temporal Alignment across Modalities},
  author={Zhou, Ziwei and Wang, Rui and Wu, Zuxuan},
  journal={arXiv preprint arXiv:2505.17862},
  year={2025}
}

@article{lian2023explainable,
  title={Explainable multimodal emotion recognition},
  author={Lian, Zheng and Sun, Haiyang and Sun, Licai and Gu, Hao and Wen, Zhuofan and Zhang, Siyuan and Chen, Shun and Xu, Mingyu and Xu, Ke and Chen, Kang and others},
  journal={arXiv preprint arXiv:2306.15401},
  year={2023}
}

@article{dong2025integrating,
  title={Integrating large language models and affective computing for human-machine symbiosis in intelligent driving},
  author={Dong, Zhiwu and Chen, Chuqiao and Liao, Chenlei and Chen, Xiqun Michael},
  journal={The Innovation},
  volume={6},
  number={12},
  year={2025},
  publisher={Elsevier}
}

@article{amin2024wide,
  title={A wide evaluation of ChatGPT on affective computing tasks},
  author={Amin, Mostafa M and Mao, Rui and Cambria, Erik and Schuller, Bj{\"o}rn W},
  journal={IEEE Transactions on Affective Computing},
  volume={15},
  number={4},
  pages={2204--2212},
  year={2024},
  publisher={IEEE}
}

@article{niu2025rethinking,
  title={Rethinking emotion annotations in the era of large language models},
  author={Niu, Minxue and El-Tawil, Yara and Romana, Amrit and Provost, Emily Mower},
  journal={IEEE Transactions on Affective Computing},
  year={2025},
  publisher={IEEE}
}

@article{yang2025large,
  title={Large language models meet text-centric multimodal sentiment analysis: A survey},
  author={Yang, Hao and Zhao, Yanyan and Wu, Yang and Wang, Shilong and Zheng, Tian and Zhang, Hongbo and Ma, Zongyang and Che, Wanxiang and Wang, Shijin and Wei, Si and others},
  journal={Science China Information Sciences},
  volume={68},
  number={10},
  pages={1--29},
  year={2025},
  publisher={Springer}
}

@incollection{cambria2017affective,
  title={Affective computing and sentiment analysis},
  author={Cambria, Erik and Das, Dipankar and Bandyopadhyay, Sivaji and Feraco, Antonio},
  booktitle={A practical guide to sentiment analysis},
  pages={1--10},
  year={2017},
  publisher={Springer}
}

@article{li2020deep,
  title={Deep facial expression recognition: A survey},
  author={Li, Shan and Deng, Weihong},
  journal={IEEE transactions on affective computing},
  volume={13},
  number={3},
  pages={1195--1215},
  year={2020},
  publisher={IEEE}
}

@article{zadeh2016multimodal,
  title={Multimodal sentiment intensity analysis in videos: Facial gestures and verbal messages},
  author={Zadeh, Amir and Zellers, Rowan and Pincus, Eli and Morency, Louis-Philippe},
  journal={IEEE Intelligent Systems},
  volume={31},
  number={6},
  pages={82--88},
  year={2016},
  publisher={IEEE}
}

@inproceedings{hazarika2020misa,
  title={Misa: Modality-invariant and-specific representations for multimodal sentiment analysis},
  author={Hazarika, Devamanyu and Zimmermann, Roger and Poria, Soujanya},
  booktitle={Proceedings of the 28th ACM international conference on multimedia},
  pages={1122--1131},
  year={2020}
}

@inproceedings{hu2022unimse,
    title = "{U}ni{MSE}: Towards Unified Multimodal Sentiment Analysis and Emotion Recognition",
    author = "Hu, Guimin  and
      Lin, Ting-En  and
      Zhao, Yi  and
      Lu, Guangming  and
      Wu, Yuchuan  and
      Li, Yongbin",
    editor = "Goldberg, Yoav  and
      Kozareva, Zornitsa  and
      Zhang, Yue",
    booktitle = "Proceedings of the 2022 Conference on Empirical Methods in Natural Language Processing",
    month = dec,
    year = "2022",
    address = "Abu Dhabi, United Arab Emirates",
    publisher = "Association for Computational Linguistics",
    url = "https://aclanthology.org/2022.emnlp-main.534/",
    doi = "10.18653/v1/2022.emnlp-main.534",
    pages = "7837--7851"
}

@article{hu2024recent,
  title={Recent trends of multimodal affective computing: A survey from NLP perspective},
  author={Hu, Guimin and Xin, Yi and Lyu, Weimin and Huang, Haojian and Sun, Chang and Zhu, Zhihong and Gui, Lin and Cai, Ruichu and Cambria, Erik and Seifi, Hasti},
  journal={arXiv preprint arXiv:2409.07388},
  year={2024}
}

@inproceedings{
lian2025affectgpt,
title={Affect{GPT}: A New Dataset, Model, and Benchmark for Emotion Understanding with Multimodal Large Language Models},
author={Zheng Lian and Haoyu Chen and Lan Chen and Haiyang Sun and Licai Sun and Yong Ren and Zebang Cheng and Bin Liu and Rui Liu and Xiaojiang Peng and Jiangyan Yi and Jianhua Tao},
booktitle={Forty-second International Conference on Machine Learning},
year={2025},
url={https://openreview.net/forum?id=xmbdACI0xu}
}

@article{zheng2023judging,
  title={Judging llm-as-a-judge with mt-bench and chatbot arena},
  author={Zheng, Lianmin and Chiang, Wei-Lin and Sheng, Ying and Zhuang, Siyuan and Wu, Zhanghao and Zhuang, Yonghao and Lin, Zi and Li, Zhuohan and Li, Dacheng and Xing, Eric and others},
  journal={Advances in neural information processing systems},
  volume={36},
  pages={46595--46623},
  year={2023}
}

@inproceedings{wang2024large,
  title={Large language models are not fair evaluators},
  author={Wang, Peiyi and Li, Lei and Chen, Liang and Cai, Zefan and Zhu, Dawei and Lin, Binghuai and Cao, Yunbo and Kong, Lingpeng and Liu, Qi and Liu, Tianyu and others},
  booktitle={Proceedings of the 62nd Annual Meeting of the Association for Computational Linguistics (Volume 1: Long Papers)},
  pages={9440--9450},
  year={2024}
}

@article{shao2024deepseekmath,
  title={Deepseekmath: Pushing the limits of mathematical reasoning in open language models},
  author={Shao, Zhihong and Wang, Peiyi and Zhu, Qihao and Xu, Runxin and Song, Junxiao and Bi, Xiao and Zhang, Haowei and Zhang, Mingchuan and Li, YK and Wu, Yang and others},
  journal={arXiv preprint arXiv:2402.03300},
  year={2024}
}

@article{lai2025rvisa,
  title={Rvisa: reasoning and verification for implicit sentiment analysis},
  author={Lai, Wenna and Xie, Haoran and Xu, Guandong and Li, Qing},
  journal={IEEE Transactions on Affective Computing},
  volume={16},
  number={3},
  pages={1760--1771},
  year={2025},
  publisher={IEEE}
}

@article{yang2025application,
  title={Application of Multiple Chain-of-Thought in Contrastive Reasoning for Implicit Sentiment Analysis},
  author={Yang, Liwei and Wang, Xinying and Zhou, Xiaotang and Wu, Zhengchao and Tan, Ningning},
  journal={arXiv preprint arXiv:2503.07140},
  year={2025}
}

@inproceedings{
wang2025videorft,
title={Video{RFT}: Incentivizing Video Reasoning Capability in {MLLM}s via Reinforced Fine-Tuning},
author={Qi Wang and Yanrui Yu and Ye Yuan and Rui Mao and Tianfei Zhou},
booktitle={The Thirty-ninth Annual Conference on Neural Information Processing Systems},
year={2025},
url={https://openreview.net/forum?id=3pORFyKzh1}
}

@article{li2025videochat,   title={Videochat-r1: Enhancing spatio-temporal perception via reinforcement fine-tuning},   author={Li, Xinhao and Yan, Ziang and Meng, Desen and Dong, Lu and Zeng, Xiangyu and He, Yinan and Wang, Yali and Qiao, Yu and Wang, Yi and Wang, Limin},   journal={arXiv preprint arXiv:2504.06958},   year={2025} }

@article{zhong2025omni,   title={Omni-R1: Reinforcement Learning for Omnimodal Reasoning via Two-System Collaboration},   author={Zhong, Hao and Zhu, Muzhi and Du, Zongze and Huang, Zheng and Zhao, Canyu and Liu, Mingyu and Wang, Wen and Chen, Hao and Shen, Chunhua},   journal={arXiv preprint arXiv:2505.20256},   year={2025} }

@article{xing2025echoink,   title={Echoink-r1: Exploring audio-visual reasoning in multimodal llms via reinforcement learning},   author={Xing, Zhenghao and Hu, Xiaowei and Fu, Chi-Wing and Wang, Wenhai and Dai, Jifeng and Heng, Pheng-Ann},   journal={arXiv preprint arXiv:2505.04623},   year={2025} }

@article{wang2025sightsound,   title={SightSound-R1: Cross-Modal Reasoning Distillation from Vision to Audio Language Models},   author={Wang, Qiaolin and Jiang, Xilin and He, Linyang and Wu, Junkai and Mesgarani, Nima},   journal={arXiv preprint arXiv:2509.15661},   year={2025} }

@article{guo2025observe,   title={Observe-r1: Unlocking reasoning abilities of mllms with dynamic progressive reinforcement learning},   author={Guo, Zirun and Hong, Minjie and Jin, Tao},   journal={arXiv preprint arXiv:2505.12432},   year={2025} }

@article{zhu2025active,   title={Active-O3: Empowering Multimodal Large Language Models with Active Perception via GRPO},   author={Zhu, Muzhi and Zhong, Hao and Zhao, Canyu and Du, Zongze and Huang, Zheng and Liu, Mingyu and Chen, Hao and Zou, Cheng and Chen, Jingdong and Yang, Ming and others},   journal={arXiv preprint arXiv:2505.21457},   year={2025} }

@misc{ chen2026grpocare, title={{GRPO}-{CARE}: Consistency-Aware Reinforcement Learning for Multimodal Reasoning}, author={Yi Chen and Yuying Ge and Rui Wang and Yixiao Ge and Junhao Cheng and Ying Shan and Xihui Liu}, year={2026}, url={https://openreview.net/forum?id=XoUJk0aDCN} }

@article{chen2025unveiling,   title={Unveiling Chain of Step Reasoning for Vision-Language Models with Fine-grained Rewards},   author={Chen, Honghao and Lou, Xingzhou and Feng, Xiaokun and Huang, Kaiqi and Wang, Xinlong},   journal={arXiv preprint arXiv:2509.19003},   year={2025} }

@article{wang2025skywork,   title={Skywork r1v2: Multimodal hybrid reinforcement learning for reasoning},   author={Wang, Peiyu and Wei, Yichen and Peng, Yi and Wang, Xiaokun and Qiu, Weijie and Shen, Wei and Xie, Tianyidan and Pei, Jiangbo and Zhang, Jianhao and Hao, Yunzhuo and others},   journal={arXiv preprint arXiv:2504.16656},   year={2025} }

@article{hong2025apo,   title={APO: Enhancing Reasoning Ability of MLLMs via Asymmetric Policy Optimization},   author={Hong, Minjie and Guo, Zirun and Xia, Yan and Wang, Zehan and Zhang, Ziang and Jin, Tao and Zhao, Zhou},   journal={arXiv preprint arXiv:2506.21655},   year={2025} }

@article{zhang2025r1,   title={R1-vl: Learning to reason with multimodal large language models via step-wise group relative policy optimization},   author={Zhang, Jingyi and Huang, Jiaxing and Yao, Huanjin and Liu, Shunyu and Zhang, Xikun and Lu, Shijian and Tao, Dacheng},   journal={arXiv preprint arXiv:2503.12937},   year={2025} }

@inproceedings{plummer2015flickr30k,   title={Flickr30k entities: Collecting region-to-phrase correspondences for richer image-to-sentence models},   author={Plummer, Bryan A and Wang, Liwei and Cervantes, Chris M and Caicedo, Juan C and Hockenmaier, Julia and Lazebnik, Svetlana},   booktitle={Proceedings of the IEEE international conference on computer vision},   pages={2641--2649},   year={2015} }

@inproceedings{nguyen2025localizing,   title={Localizing Before Answering: A Benchmark for Grounded Medical Visual Question Answering},   author={Nguyen, Dung and Ho, Minh Khoi and Ta, Huy and Nguyen, Thanh Tam and Chen, Qi and Rav, Kumar and Dang, Quy Duong and Ramchandre, Satwik and Phung, Son Lam and Liao, Zhibin and others},   booktitle={Proceedings of the Thirty-Fourth International Joint Conference on Artificial Intelligence},   pages={7670--7678},   year={2025} }

@inproceedings{chen2022grounding,   title={Grounding answers for visual questions asked by visually impaired people},   author={Chen, Chongyan and Anjum, Samreen and Gurari, Danna},   booktitle={Proceedings of the IEEE/CVF Conference on Computer Vision and Pattern Recognition},   pages={19098--19107},   year={2022} }

@inproceedings{cekinel2025multimodal,   title={Multimodal fact-checking with vision language models: A probing classifier based solution with embedding strategies},   author={Cekinel, Recep Firat and Karagoz, Pinar and {\c{C}}{\"o}ltekin, {\c{C}}a{\u{g}}r{\i}},   booktitle={Proceedings of the 31st International Conference on Computational Linguistics},   pages={4622--4633},   year={2025} }

@inproceedings{yu2024rlhf,   title={Rlhf-v: Towards trustworthy mllms via behavior alignment from fine-grained correctional human feedback},   author={Yu, Tianyu and Yao, Yuan and Zhang, Haoye and He, Taiwen and Han, Yifeng and Cui, Ganqu and Hu, Jinyi and Liu, Zhiyuan and Zheng, Hai-Tao and Sun, Maosong and others},   booktitle={Proceedings of the IEEE/CVF Conference on Computer Vision and Pattern Recognition},   pages={13807--13816},   year={2024} }

@inproceedings{sun2024aligning,   title={Aligning large multimodal models with factually augmented rlhf},   author={Sun, Zhiqing and Shen, Sheng and Cao, Shengcao and Liu, Haotian and Li, Chunyuan and Shen, Yikang and Gan, Chuang and Gui, Liangyan and Wang, Yu-Xiong and Yang, Yiming and others},   booktitle={Findings of the Association for Computational Linguistics: ACL 2024},   pages={13088--13110},   year={2024} }

@article{zhang2025improving,   title={Improving the reasoning of multi-image grounding in mllms via reinforcement learning},   author={Zhang, Bob and Li, Haoran and Zhang, Tao and Yan, Cilin and Cai, Jiayin and Hao, Yanbin},   journal={arXiv preprint arXiv:2507.00748},   year={2025} }

@article{feng2025video,
  title={Video-R1: Reinforcing Video Reasoning in MLLMs},
  author={Feng, Kaituo and Gong, Kaixiong and Li, Bohao and Guo, Zonghao and Wang, Yibing and Peng, Tianshuo and Wang, Benyou and Yue, Xiangyu},
  journal={arXiv preprint arXiv:2503.21776},
  year={2025}
}

@misc{siq2,
  author = {Alex Wilf and Leena Mathur and Sheryl Mathew and Claire Ko and Youssouf Kebe and Paul Pu Liang and Louis-Philippe Morency},
  title = {Social-IQ 2.0 Challenge: Benchmarking Multimodal Social Understanding},
  year = {2023},
  publisher = {GitHub},
  journal = {GitHub repository},
  howpublished = {\url{https://github.com/abwilf/Social-IQ-2.0-Challenge}},
}

@article{hong2025worldsense,
  title={Worldsense: Evaluating real-world omnimodal understanding for multimodal llms},
  author={Hong, Jack and Yan, Shilin and Cai, Jiayin and Jiang, Xiaolong and Hu, Yao and Xie, Weidi},
  journal={arXiv preprint arXiv:2502.04326},
  year={2025}
}

@article{hurst2024gpt,
  title={Gpt-4o system card},
  author={Hurst, Aaron and Lerer, Adam and Goucher, Adam P and Perelman, Adam and Ramesh, Aditya and Clark, Aidan and Ostrow, AJ and Welihinda, Akila and Hayes, Alan and Radford, Alec and others},
  journal={arXiv preprint arXiv:2410.21276},
  year={2024}
}

@article{liu2025ola,
  title={Ola: Pushing the frontiers of omni-modal language model},
  author={Liu, Zuyan and Dong, Yuhao and Wang, Jiahui and Liu, Ziwei and Hu, Winston and Lu, Jiwen and Rao, Yongming},
  journal={arXiv preprint arXiv:2502.04328},
  year={2025}
}

@article{fu2025vita,
  title={Vita-1.5: Towards gpt-4o level real-time vision and speech interaction},
  author={Fu, Chaoyou and Lin, Haojia and Wang, Xiong and Zhang, Yi-Fan and Shen, Yunhang and Liu, Xiaoyu and Cao, Haoyu and Long, Zuwei and Gao, Heting and Li, Ke and others},
  journal={arXiv preprint arXiv:2501.01957},
  year={2025}
}

@article{yao2024minicpm,
  title={Minicpm-v: A gpt-4v level mllm on your phone},
  author={Yao, Yuan and Yu, Tianyu and Zhang, Ao and Wang, Chongyi and Cui, Junbo and Zhu, Hongji and Cai, Tianchi and Li, Haoyu and Zhao, Weilin and He, Zhihui and others},
  journal={arXiv preprint arXiv:2408.01800},
  year={2024}
}

@article{jaech2024openai,
  title={Openai o1 system card},
  author={Jaech, Aaron and Kalai, Adam and Lerer, Adam and Richardson, Adam and El-Kishky, Ahmed and Low, Aiden and Helyar, Alec and Madry, Aleksander and Beutel, Alex and Carney, Alex and others},
  journal={arXiv preprint arXiv:2412.16720},
  year={2024}
}

@misc{google2025gemini25propreview,
  author = {{Google DeepMind}},
  title  = {{Gemini 2.5 Pro Preview}: even better coding performance},
  year   = {2025},
  note   = {Blog post. Available at: https://deepmind.google/blog/gemini-25-pro-preview-even-better-coding-performance/ (accessed 2026-01-21).}
}

@inproceedings{lu2024unified,
  title={Unified-io 2: Scaling autoregressive multimodal models with vision language audio and action},
  author={Lu, Jiasen and Clark, Christopher and Lee, Sangho and Zhang, Zichen and Khosla, Savya and Marten, Ryan and Hoiem, Derek and Kembhavi, Aniruddha},
  booktitle={Proceedings of the IEEE/CVF Conference on Computer Vision and Pattern Recognition},
  pages={26439--26455},
  year={2024}
}

@article{cheng2024videollama,
  title={Videollama 2: Advancing spatial-temporal modeling and audio understanding in video-llms},
  author={Cheng, Zesen and Leng, Sicong and Zhang, Hang and Xin, Yifei and Li, Xin and Chen, Guanzheng and Zhu, Yongxin and Zhang, Wenqi and Luo, Ziyang and Zhao, Deli and others},
  journal={arXiv preprint arXiv:2406.07476},
  year={2024}
}

@misc{anthropic2024claude3family,
  author       = {{Anthropic}},
  title        = {Introducing the next generation of Claude},
  year         = {2024},
  url          = {https://www.anthropic.com/news/claude-3-family},
  urldate      = {2024-10-22},
  note         = {Accessed: 2024-10-22}
}

@article{team2024gemini,
  title={Gemini 1.5: Unlocking multimodal understanding across millions of tokens of context},
  author={Team, Gemini and Georgiev, Petko and Lei, Ving Ian and Burnell, Ryan and Bai, Libin and Gulati, Anmol and Tanzer, Garrett and Vincent, Damien and Pan, Zhufeng and Wang, Shibo and others},
  journal={arXiv preprint arXiv:2403.05530},
  year={2024}
}

@inproceedings{xiao2021next,
  title={Next-qa: Next phase of question-answering to explaining temporal actions},
  author={Xiao, Junbin and Shang, Xindi and Yao, Angela and Chua, Tat-Seng},
  booktitle={Proceedings of the IEEE/CVF conference on computer vision and pattern recognition},
  pages={9777--9786},
  year={2021}
}

@article{zhang2024llava,
  title={Llava-video: Video instruction tuning with synthetic data},
  author={Zhang, Yuanhan and Wu, Jinming and Li, Wei and Li, Bo and Ma, Zejun and Liu, Ziwei and Li, Chunyuan},
  journal={arXiv preprint arXiv:2410.02713},
  year={2024}
}

@article{xu2024pllava,
  title={Pllava: Parameter-free llava extension from images to videos for video dense captioning},
  author={Xu, Lin and Zhao, Yilin and Zhou, Daquan and Lin, Zhijie and Ng, See Kiong and Feng, Jiashi},
  journal={arXiv preprint arXiv:2404.16994},
  year={2024}
}

@inproceedings{yang2025magma,
  title={Magma: A foundation model for multimodal ai agents},
  author={Yang, Jianwei and Tan, Reuben and Wu, Qianhui and Zheng, Ruijie and Peng, Baolin and Liang, Yongyuan and Gu, Yu and Cai, Mu and Ye, Seonghyeon and Jang, Joel and others},
  booktitle={Proceedings of the computer vision and pattern recognition conference},
  pages={14203--14214},
  year={2025}
}

@inproceedings{li2024mvbench,
  title={Mvbench: A comprehensive multi-modal video understanding benchmark},
  author={Li, Kunchang and Wang, Yali and He, Yinan and Li, Yizhuo and Wang, Yi and Liu, Yi and Wang, Zun and Xu, Jilan and Chen, Guo and Luo, Ping and others},
  booktitle={Proceedings of the IEEE/CVF Conference on Computer Vision and Pattern Recognition},
  pages={22195--22206},
  year={2024}
}

@article{li2024llava,
  title={Llava-onevision: Easy visual task transfer},
  author={Li, Bo and Zhang, Yuanhan and Guo, Dong and Zhang, Renrui and Li, Feng and Zhang, Hao and Zhang, Kaichen and Zhang, Peiyuan and Li, Yanwei and Liu, Ziwei and others},
  journal={arXiv preprint arXiv:2408.03326},
  year={2024}
}

@article{ye2024mplug,
  title={mplug-owl3: Towards long image-sequence understanding in multi-modal large language models},
  author={Ye, Jiabo and Xu, Haiyang and Liu, Haowei and Hu, Anwen and Yan, Ming and Qian, Qi and Zhang, Ji and Huang, Fei and Zhou, Jingren},
  journal={arXiv preprint arXiv:2408.04840},
  year={2024}
}

\section{Biography Section}
\vspace{-4em}
\newcommand{\bioimg}[1]{\includegraphics[width=0.75in,height=0.94in,clip,keepaspectratio]{#1}}
\newcommand{\biogap}{\vspace*{-3em}}

\begin{IEEEbiography}[{\includegraphics[width=1in,height=1.25in,clip,keepaspectratio]{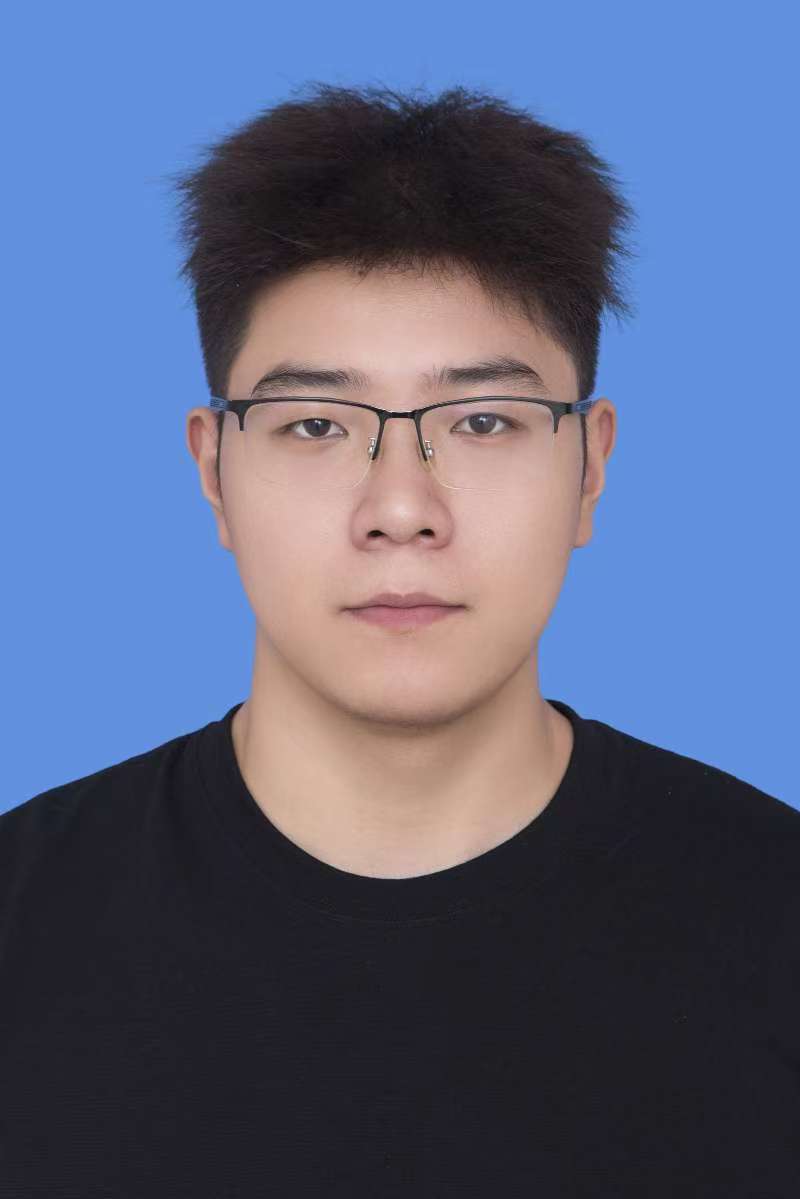}}]{Yibo Wang}
is currently pursuing the M.S. degree at the Gansu Provincial
Key Laboratory of Wearable Computing, School of Information
Science and Engineering, Lanzhou University, Lanzhou, China.
He received the B.S. degree from Dalian University of
Technology, Dalian, China. His research interests include
multimodal large language models, affective computing, and
reinforcement learning.
\end{IEEEbiography}
\biogap

\begin{IEEEbiography}[{\includegraphics[width=1in,height=1.25in,clip,keepaspectratio]{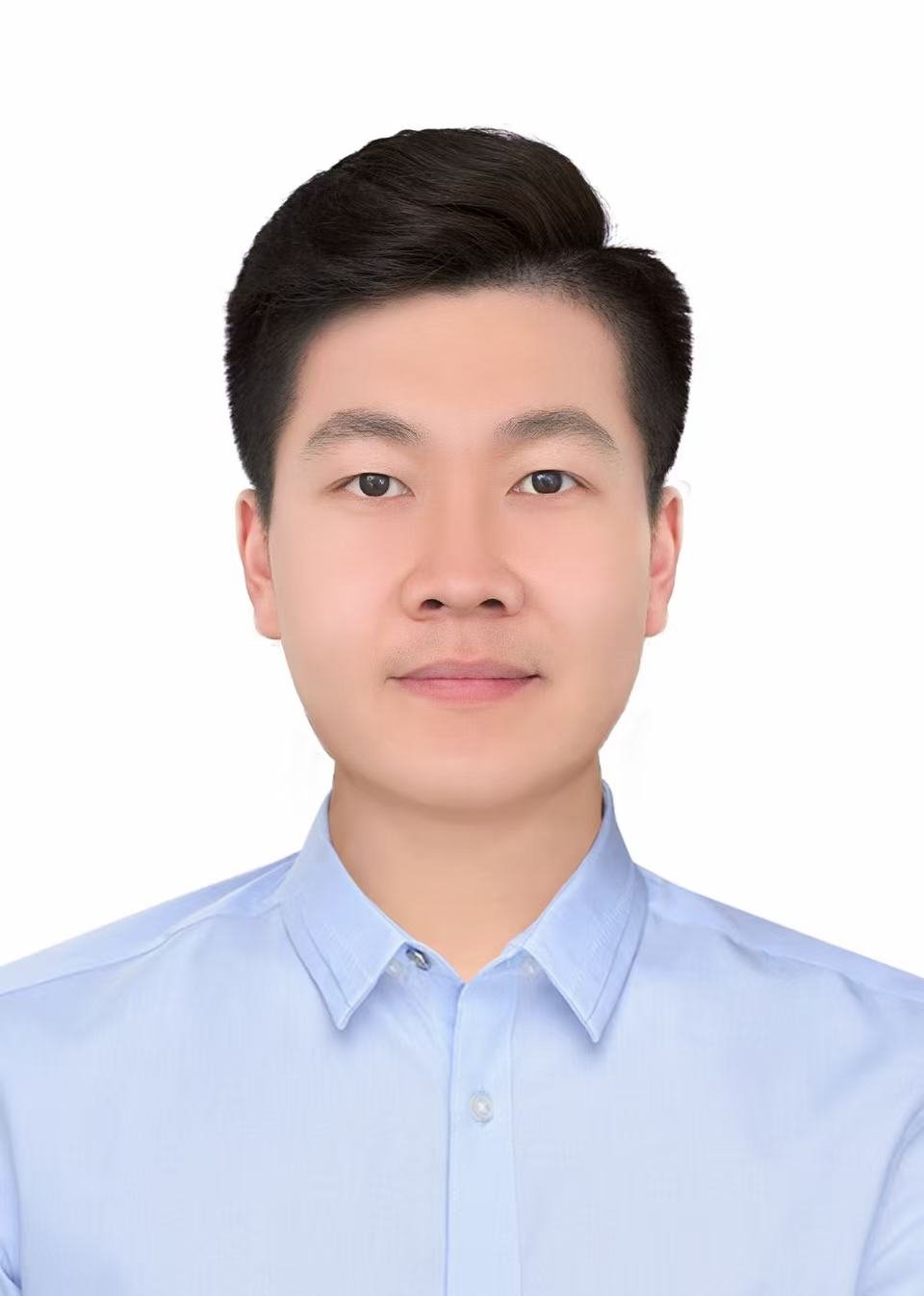}}]{Rui Yang}
is currently pursuing the Ph.D. degree with the Gansu Provincial Key Laboratory of Wearable Computing, School of Information Science and Engineering, Lanzhou University. His research interests include biometric authentication, affective computing, large language models (LLMs), and embodied intelligence.
\end{IEEEbiography}

\begin{IEEEbiography}[{\includegraphics[width=1in,height=1.25in,clip,keepaspectratio]{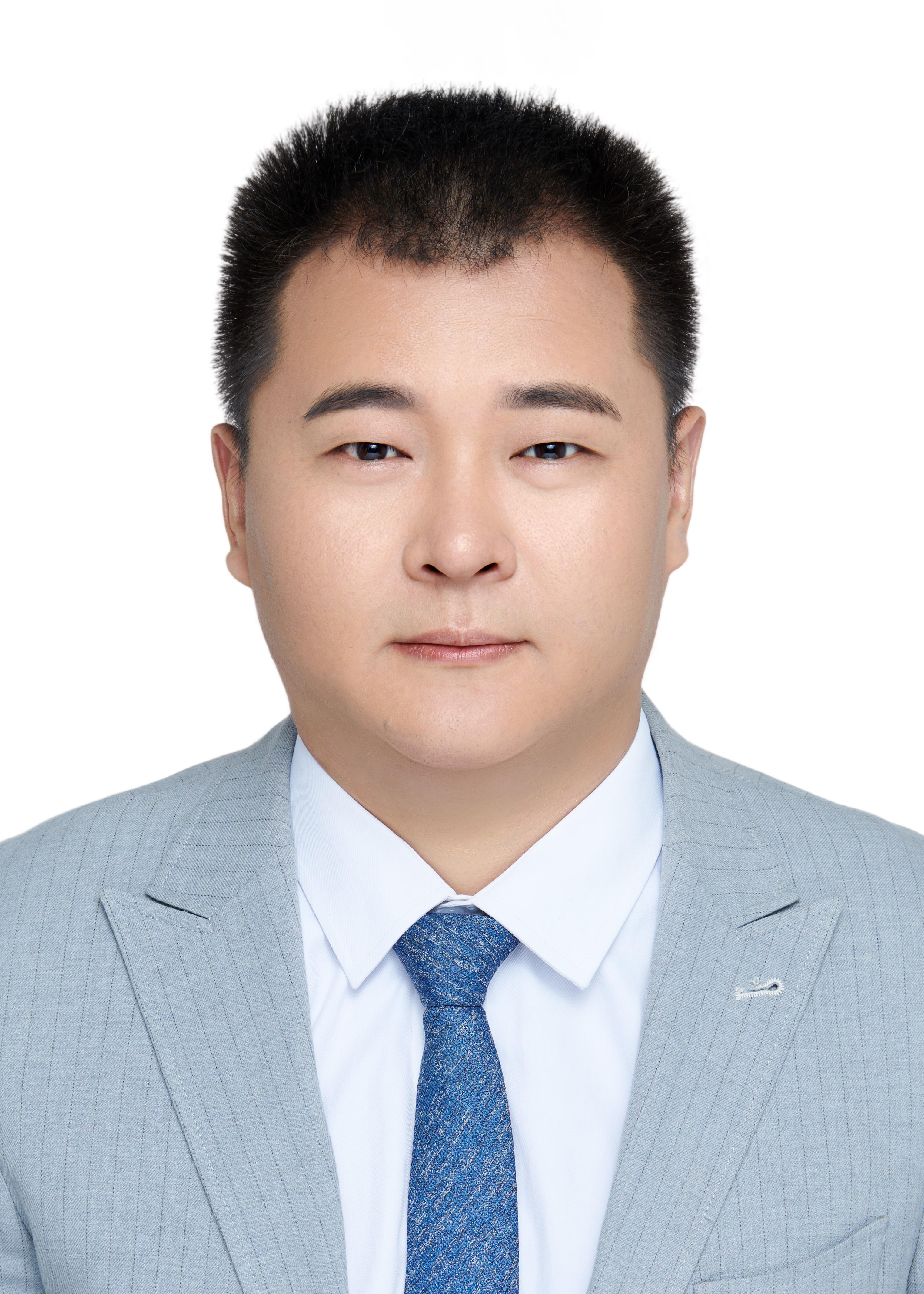}}]{Jisheng Dang}
received the Ph.D. degree from Sun Yat-sen University, China,
in 2025, advised by prof. Jianhuang Lai and prof. Huicheng
Zheng. He worked as a research fellow at the NExT++ laboratory
of the National University of Singapore advised by prof.
Tat-Seng Chua. He is now a tenured associate professor at the
School of Information Science and Engineering, Lanzhou
University. His research interests include multimodal learning,
video understanding, and embodied intelligence. He has
published several papers as the first author in major journals
and conferences including IEEE TIP/TNNLS/TITS/IJCAI/AAAI.
He served as a reviewer at some major journals and conferences
like IEEE TPAMI, ICML, NIPS, ICLR, IEEE TIP, CVPR, IJCAI,
ACM MM, AAAI, IEEE TMM, IEEE TCSVT, ACM TOMM.
\end{IEEEbiography}
\biogap

\begin{IEEEbiography}[{\includegraphics[width=1in,height=1.25in,clip,keepaspectratio]{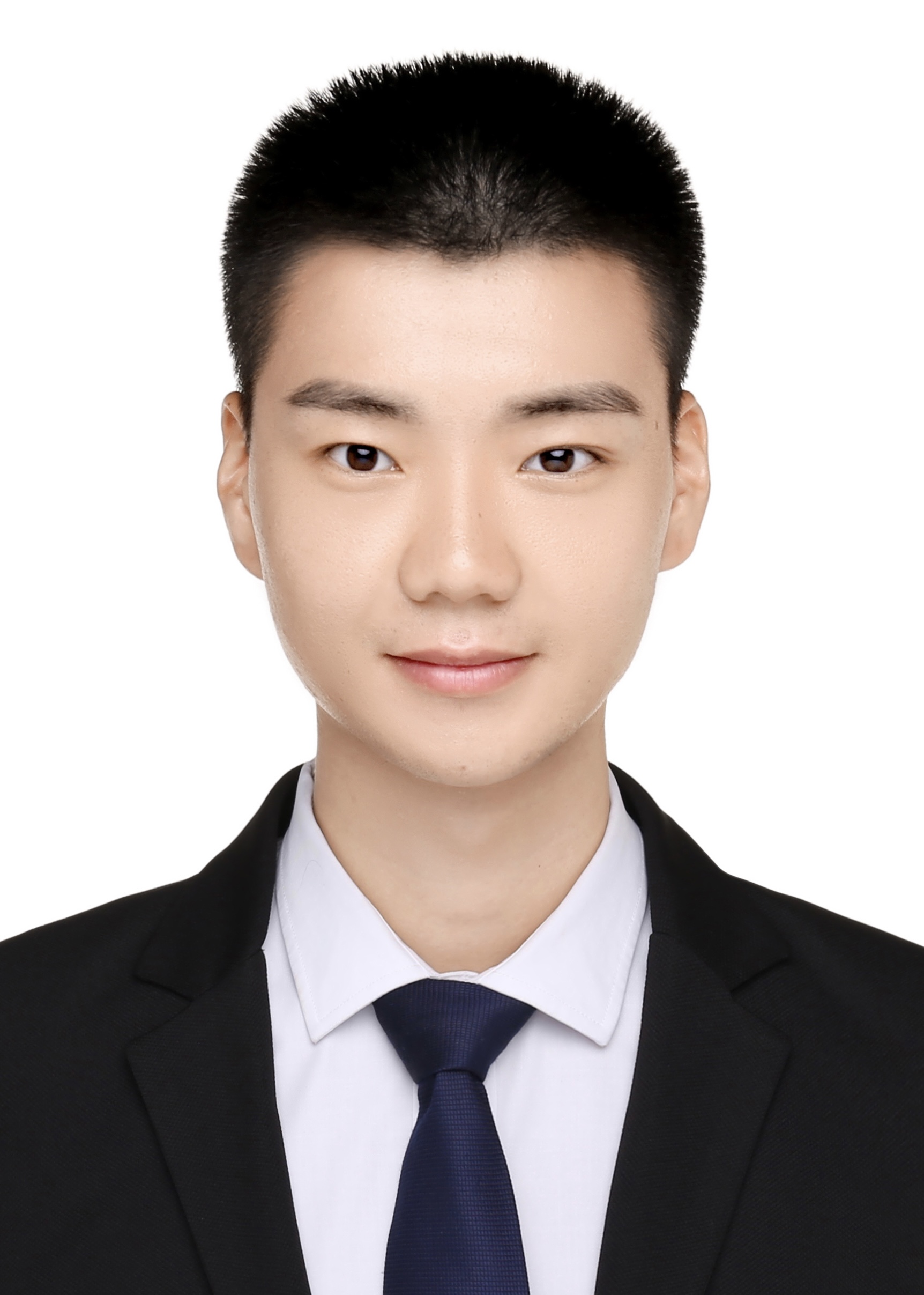}}]{Yitao Wu}
received the B.S. degree in information system and information management from Hainan University. His research interests include large language models, vision-language-action models, and vision-language models.
\end{IEEEbiography}

\begin{IEEEbiography}[{\includegraphics[width=1in,height=1.25in,clip,keepaspectratio]{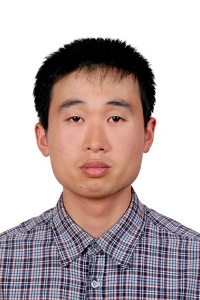}}]{Hong Peng}
received the Ph.D. degree from Lanzhou University, Lanzhou,
China. From 2010 to 2011, he was a Visiting Scholar with the
Institute of Computer System, ETH Zurich, Switzerland. He is
currently an Associate Professor with the School of Information
Science and Engineering, Lanzhou University. He is also in
charge of three projects from the National Natural Science
Foundation of China, the Central College Foundation Project of
Lanzhou University, and the Youth Cross-Project of Lanzhou
University. He has authored or coauthored more than 30 papers
in peer-reviewed journals, conferences, and book chapters.
His research areas include bioinformation processing and
ubiquitous affective computing.
\end{IEEEbiography}
\biogap

\begin{IEEEbiography}[{\includegraphics[width=1in,height=1.25in,clip,keepaspectratio]{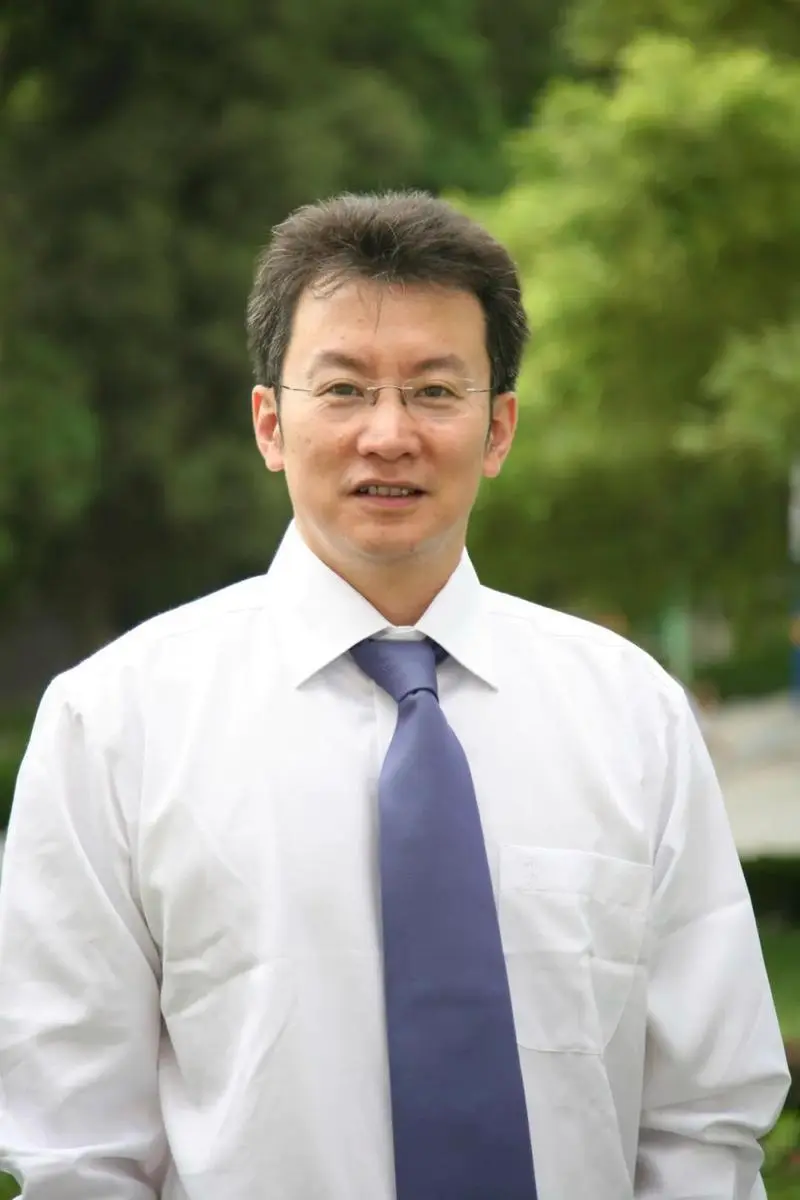}}]{Bin Hu}
(Fellow, IEEE) received the PhD degree in computer science
from the Institute of Computing Technology, Chinese Academy
of Science, China, in 1998. Since 2008, he has been a
professor and the dean of the School of Information Science
and Engineering, Lanzhou University, China. He had been also
guest professorship in ETH Zurich, Switzerland till 2011.
He is a Professor of Lanzhou University and Beijing Institute
of Technology. He serves as Editor-in-Chief of IEEE
Transactions on Computational Social Systems, Fellow of IET
and AAIA, and Chair of Technical Committee on Computational
Psychophysiology, IEEE SMC. He has published over 300 papers
in domestic and international academic journals and
conferences. His research interests include pervasive
computing, computational psychophysiology, data modeling,
and artificial intelligence.
\end{IEEEbiography}
\biogap

\begin{IEEEbiography}[{\includegraphics[width=1in,height=1.25in,clip,keepaspectratio]{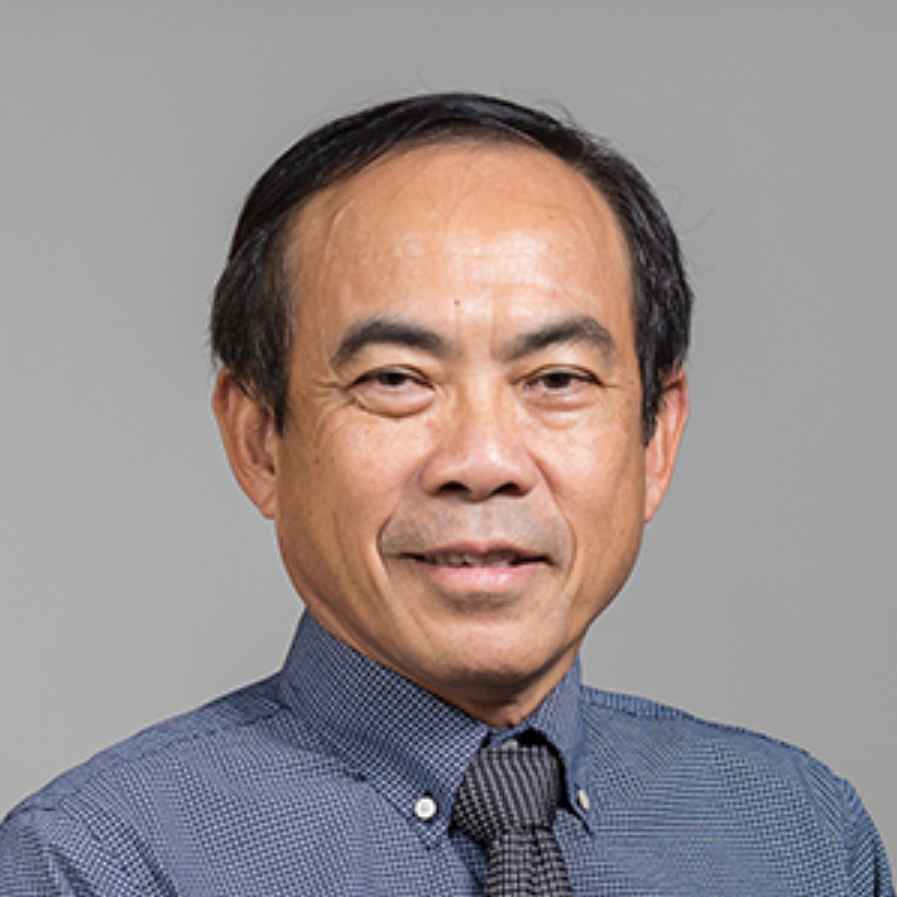}}]{Tat-Seng Chua}
received the Ph.D. degree from the University of Leeds, U.K.
He is the KITHCT chair professor with the School of Computing,
National University of Singapore, where he was the acting and
founding dean of the School from 1998 to 2000. He is the
co-director of NExT, a joint center between NUS and Tsinghua
University, to develop technologies for live social media
search. He is the 2015 winner of the prestigious ACM SIGMM
Award. He is the chair of Steering Committee of the ACM
International Conference on Multimedia Retrieval (ICMR) and
Multimedia Modeling (MMM) conference series. He is also the
general co-chair of ACM Multimedia 2005, ACM CIVR (now ACM
ICMR) 2005, ACM SIGIR 2008, and ACM Web Science 2015. He
serves on the editorial boards of four international journals.
He is the co-founder of two technology startups in Singapore
and a Fellow of the Singapore Academy of Sciences, with 107,618 citations on Google Scholar.
\end{IEEEbiography}

\vfill

\end{document}